\documentclass[letterpaper, 10 pt, conference]{ieeeconf}
\usepackage[utf8]{inputenc}
\usepackage{amsmath}
\usepackage{amsfonts}
\usepackage{amssymb}
\usepackage{subcaption}
\usepackage{bm}
\usepackage{url}
\usepackage{graphicx}

\graphicspath{{./pictures/}}

\usepackage[bottom]{footmisc}

\IEEEoverridecommandlockouts                              
\title{\LARGE \bf
Comparison of Motion Encoding Frameworks on \\Human Manipulation Actions
}

\author{Lennart Jahn,
        Florentin W\"org\"otter,
        and Tomas Kulvicius
\thanks{This work was supported by the the Volkswagen Foundation (``DeMoDiag'', grant number ZN3543) and by the German Science Foundation Grant (DFG WO 388/16-1).}
\thanks{L. Jahn and F. W\"org\"otter are with the Third Institute of Physics, Dept. Computational Neuroscience, University of G\"ottingen, 37073 G\"ottingen, Germany, e-mail: lennart.jahn@phys.uni-goettingen.de.}
\thanks{T. Kulvicius is with the University Medical Center G\"ottingen, Child and Adolescent Psychiatry and Psychotherapy, 37075 G\"ottingen, Germany.}
}

\begin{document}

\maketitle

\begin{abstract}
    Movement generation, and especially generalisation to unseen situations, plays an important role in robotics. Different types of movement generation methods exist such as spline based methods, dynamical system based methods, and methods based on Gaussian mixture models (GMMs). Using a large, new dataset on human manipulations, in this paper we provide a highly detailed comparison of five fundamentally different and widely used movement encoding and generation frameworks: dynamic movement primitives (DMPs), time based Gaussian mixture regression (tbGMR), stable estimator of dynamical systems (SEDS), Probabilistic Movement Primitives (ProMP) and Optimal Control Primitives (OCP). We compare these frameworks with respect to their movement encoding efficiency, reconstruction accuracy, and movement generalisation capabilities. The new dataset consists of nine object manipulation actions performed by 12 humans: pick and place, put on top/take down, put inside/take out, hide/uncover, and push/pull with a total of 7,652 movement examples. 

    Our analysis shows that for movement encoding and reconstruction DMPs and OCPs are the most efficient with respect to the number of parameters and reconstruction accuracy, if a sufficient number of kernels is used. In case of movement generalisation to new start- and end-point situations, DMPs, OCPs and task parameterized GMM (TP-GMM, movement generalisation framework based on tbGMR) lead to similar performance, which ProMPs only achieve when using many demonstrations for learning. All models outperform SEDS, which additionally proves to be difficult to fit. 
    Furthermore we observe that TP-GMM and SEDS suffer from problems reaching the end-points of generalizations.
    These different quantitative results will help selecting the most appropriate models and designing trajectory representations in an improved task-dependent way in future robotic applications. 
\end{abstract}

\section{Introduction}

Movement generation methods play an important role in robotics that is crucial for enabling robots to perform actions precisely, and especially to generalize movements to unknown situations. Apart from industrial applications, where fast and reliably repetitive movements are needed, static, non-adaptive trajectory representations such as interpolation between via-points using splines \cite{Siciliano2009} are insufficient in most cases.
In dynamic environments, trajectories need to be adapted, and it is infeasible to pre-program trajectories for each possible situation.
Therefore, motion encoding frameworks need to be able to generalize movements to new situations, e.g., to new start- or end-points.
Furthermore, this way, they are more easily allowing for learning by demonstration \cite{Billard_RobotProgrammingDemonstration_2008, Ravichandar_RecentAdvancesRobot_2020a}, also called transfer learning or imitation learning.
In imitation learning, instead of obtaining new movements by a theoretical description or optimization method, human demonstrations are encoded by some mathematical model and then used by a robot.
Human demonstrations are usually obtained by manually moving the robot, called kinesthetic guidance, or via motion tracking of the human body.

To generalize trajectories to new situations, a motion encoding framework should therefore be able to generate human like movements for new situations from given demonstrations of known situations.
As will be discussed below, various different approaches were developed for movement generalization, however the question arises, which of these frameworks performs best with respect to efficiency and accuracy of movement encoding and generalization to unseen situations.

\paragraph{Motion encoding frameworks}
Dynamical Movement Primitives (DMPs,  \cite{Ijspeert_DynamicalMovementPrimitives_2013}) are one of the most frequently used encoding frameworks.
There, a critically damped linear attractor system is forced to follow a desired trajectory by a nonlinear term that is approximated by a weighted sum of Gaussian kernels.
The model is so widely used that it has been modified and extended in numerous ways.
There have been alterations to the weight fitting process, for example to enable learning from multiple demonstrations \cite{Yin_Learningnonlineardynamical_2014, Pervez_Novellearningdemonstration_2017}, extensions to include rotational orientation \cite{Ude_OrientationCartesianspace_2014a, Abu-Dakka_Adaptationmanipulationskills_2015, Abu-Dakka_GeometryawareDynamicMovement_2020}, or improving the generalization capabilities, for example by biologically motivated changes to the DMPs force field \cite{Hoffmann_Biologicallyinspireddynamicalsystems_2009} or having via-points on the way \cite{Weitschat_SafeEfficientHuman_2018}.
There is an exhaustive review by Saveriano et al. \cite{Saveriano_Dynamicmovementprimitives_2023} on that matter.

Via-Point Movement Primitives \cite{Zhou_LearningViaPointMovement_2019} also increase the capability of the system to adapt to new situations using via-points along the trajectory, but extend the DMP model further, by introducing components of Probabilistic Movement Primitives, which will be mentioned later.

A different approach that also uses a representation of a dynamical system is Optimal Control Primitives \cite{Herzog_Generationmovementsboundary_2017}.
There, the trajectories are represented with Chebyshev polynomials instead of Gaussian kernels.
A linear-quadratic regulator, derived from optimal control theory (hence the name), ensures the accurate reproduction of the trajectories and robustness against perturbations.
All these frameworks are based on the common ground of dynamical systems with implicit time dependence, which means that the trajectories are obtained by integrating a specially designed dynamical system that then generates the desired trajectories.

Another class of approaches uses Gaussian Mixture Models (GMMs) to encode trajectories.
Since GMMs in general are a model for regression of arbitrary distributions, they are commonly used for many other applications in statistics and data science.
Moreover, they can also be used to estimate the parameters of frameworks with dynamical systems similar to the methods above \cite{Khansari-Zadeh_BMiterativealgorithm_2010, Khansari-Zadeh_LearningStableNonlinear_2011}.
One key difference to the previously mentioned methods is that GMMs are able to encode multidimensional trajectories in one model, and thus, encode correlations between the dimensions, while most dynamical systems-based approaches use as many independent 1D systems as there are dimensions in the trajectory.
GMM based models can also work completely without dynamical systems, for example by encoding the joint distribution of position and time and extracting the trajectory using the conditional distribution of position \textit{given} time \cite{Calinon_LearningRepresentingGeneralizing_2007, Calinon_tutorialtaskparameterizedmovement_2016}.
Recently, a deep learning based approach was proposed to encode GMMs using a mixture density network (MDN, \cite{9057560}).

Yet another different class of encoding frameworks, called transverse contracting dynamic system primitive (CDSP), uses contraction analysis \cite{Manchester_Transversecontractioncriteria_2014} to estimate parameters of an autonomous dynamical system and store them either in a GMM \cite{Ravichandar_Learningpositionorientation_2019} or a neural network \cite{Ravichandar_Learningperiodicmotions_2016}.

Methods like Probabilistic Movement Primitives \cite{Paraschos_ProbabilisticMovementPrimitives_2013} exist, which can encode cross-correlations, too; but their working principle of modelling distributions over weight based representations of trajectories is closer to the GMM models than to DMPs.
Recently, the TP-GMM approach used for generalizing GMMs was combined with ProMPs to further improve their generalization capabilities \cite{Yao_ImprovedGeneralizationProbabilistic_2024}.

\paragraph{Own contribution}
There exist several papers comparing different models either only theoretically, e.g., for DMPs \cite{Saveriano_Dynamicmovementprimitives_2023}, or with application on a common task, e.g., for DMPs, PMPs and OCPs \cite{Herzog_Generationmovementsboundary_2017} or for different kinds of GMM encodings \cite{Calinon_improvingextrapolationcapability_03}.
Those papers compare relatively similar frameworks on small datasets and their respective tasks are not comparable between each other.
There also exists a study on benchmarking of different models \cite{Lemme_Opensourcebenchmarkinglearned_2015} which focuses on measures to judge the human-likeness of actions generated by several different frameworks, as well as their robustness to perturbations.
This benchmarking was performed on a dataset of handwriting trajectories, however, there was no quantitative comparison of generalized trajectories (i.e., with respect to the changed end-point) to ground truth (human) trajectories.
In the current work we aim to provide a thorough analysis and comparison of different type models on a larger dataset (9 different action classes; 7,652 movement trajectories in total) with the focus on generalization of point-to-point human hand movements to new start-/end-positions that plays an important role for generation of human-like robotic manipulation actions in unknown situations.

We compare five different frameworks, namely DMPs \cite{Ijspeert_DynamicalMovementPrimitives_2013}, task parameterized GMMs (TP-GMMs) with time based Gaussian Mixture Regression (tbGMR, \cite{Calinon_tutorialtaskparameterizedmovement_2016}), stable estimator of dynamical systems (SEDS,  \cite{Khansari-Zadeh_LearningStableNonlinear_2011}), Probabilistic Movement Primitives (ProMPs, \cite{NIPS2013_e53a0a29}) and Optimal Control Primitives (OCPs, \cite{Herzog_Generationmovementsboundary_2017}).
These motion encoding frameworks have been widely used for robot motion control and demonstration learning \cite{Calinon_LearningRepresentingGeneralizing_2007, Hung_approachlearnhand_2016, Ude_TaskSpecificGeneralizationDiscrete_2010a, Billard_RobotProgrammingDemonstration_2008, Ravichandar_RecentAdvancesRobot_2020a, Paraschos_probabilisticapproachrobot_2015, Paraschos_Usingprobabilisticmovement_2018}.

We selected those five models, because they represent a wide variety of different encoding methods.
These methods are either based on dynamical systems (DMPs, SEDS, OCPs) or not (GMMs, ProMPs), use different dimensions for movement encoding and a different number of encoding parameters, can encode single or multiple trajectories in one model, and generalize in different ways.
We are aware, that there exist multiple extensions and variations of these models as cited above, but in this work we compare the general basic models on a rather general movement encoding and generalization task.
In some specific cases and applications, a more specialized sub-type of those models may lead to better performance.

We quantitatively analyze and compare the performance of the frameworks for movement encoding and reconstruction, and for generalization from demonstrations to new situations on a novel very large dataset of human manipulation actions. The main novel aspect compared to the previous work and the key focus of this study is on how well different movement generation frameworks are able to generalise to new situations in comparison to human movements.  

First, we investigate how well the frameworks are able to encode and reconstruct single trajectories from the action dataset. Second, we also test the generalization capabilities of the frameworks by presenting subsets of the dataset as training demonstrations and testing the generalization capabilities on different situations obtained from the remaining demonstrations. Here we evaluate how similar trajectories generated by different models are to the human movements.
Furthermore, we also compare how many free parameters the models need to achieve their respective performance.

\section{Dataset}

\subsection{Recording setup}
\begin{figure}[ht]
	\centering
	\begin{subfigure}[t]{0.95\linewidth}
		\includegraphics[width=\textwidth]{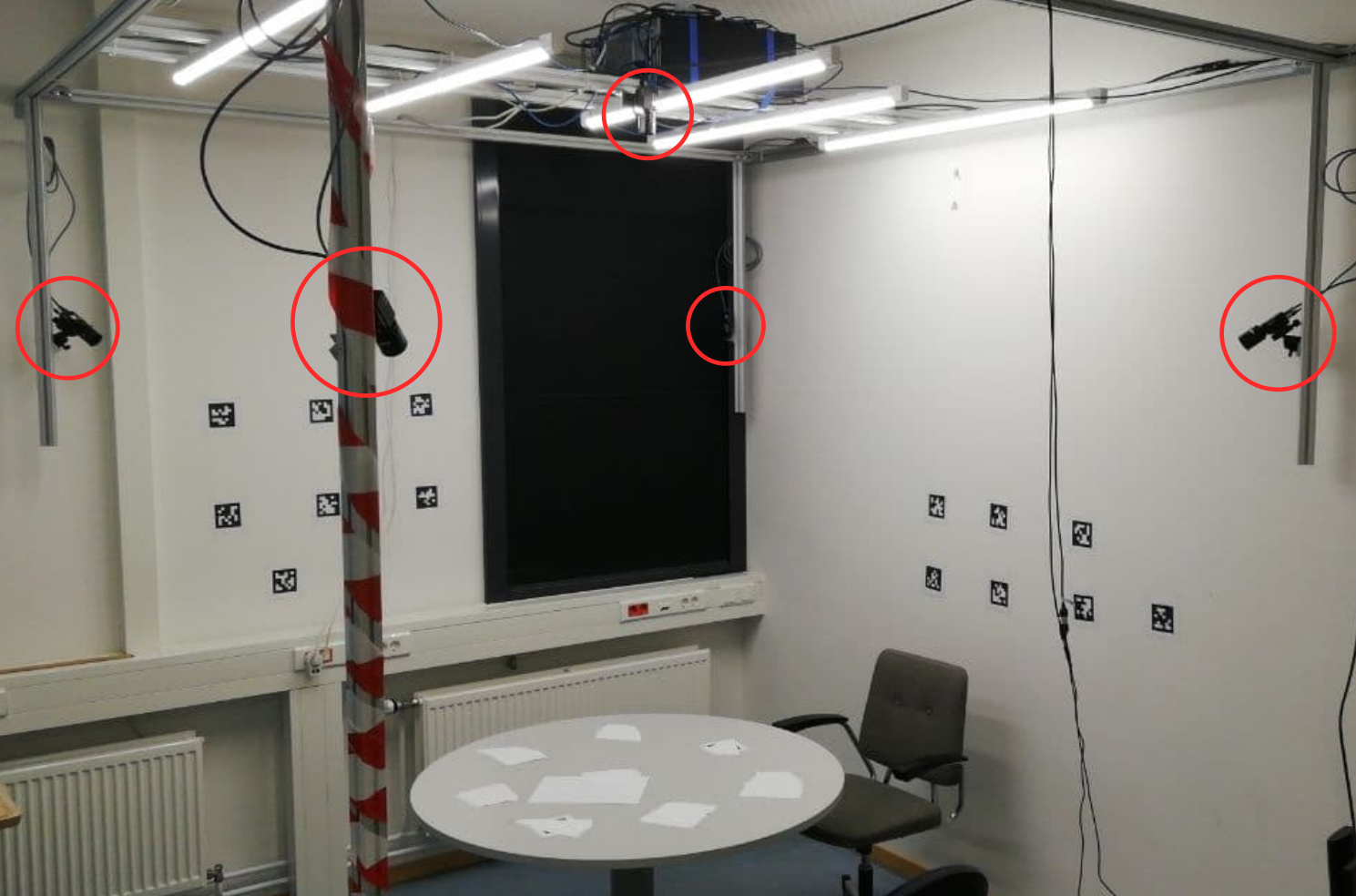}
		\subcaption{The multi-camera recording setup with one top-view and four side-view cameras (marked by red circles).}
		\label{fig:setupIRL}
	\end{subfigure}
	\begin{subfigure}[t]{0.45\linewidth}
		\includegraphics[width=\textwidth]{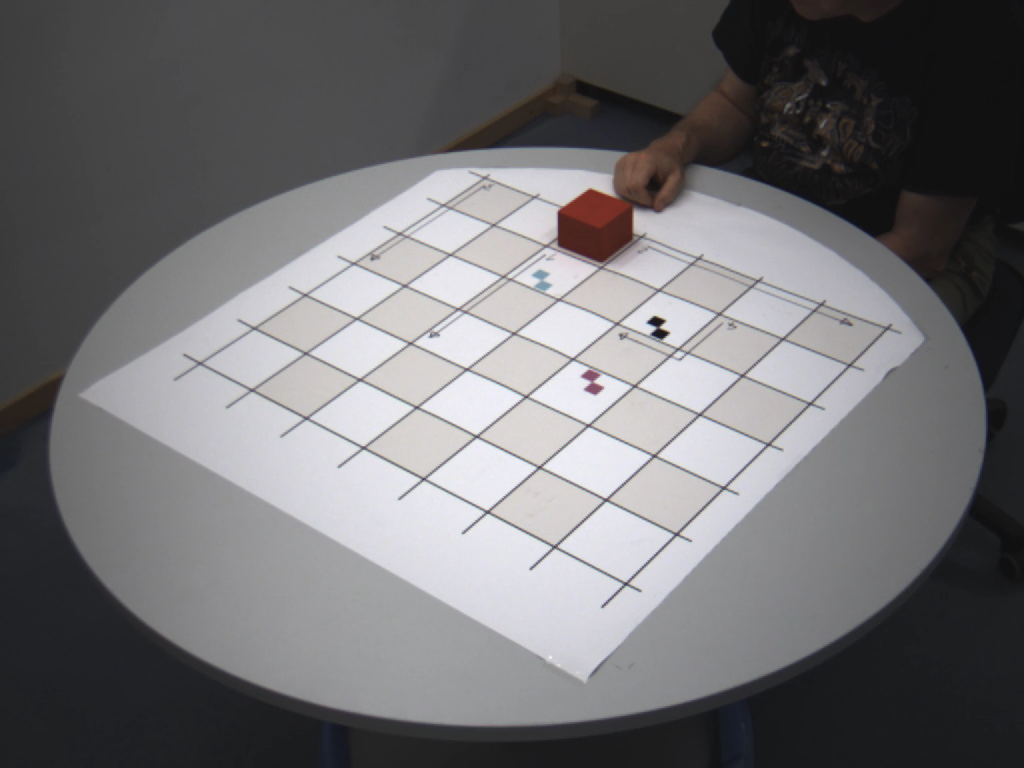}
		\subcaption{Table setup view with the checkerboard mat.}
		\label{fig:exampleview}
	\end{subfigure}
	\begin{subfigure}[t]{0.45\linewidth}
		\includegraphics[width=\textwidth]{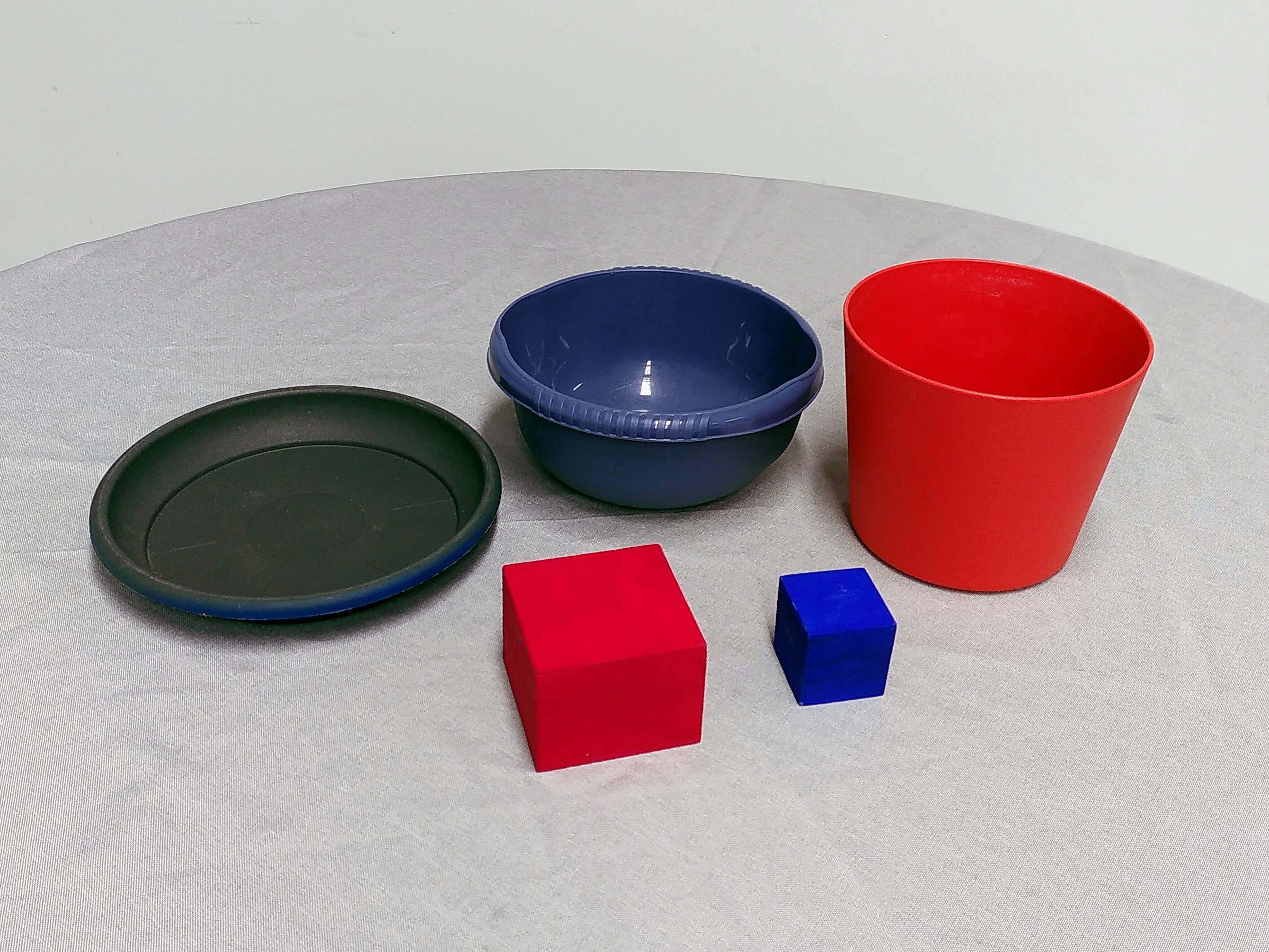}
		\subcaption{Objects used for the recordings.}
		\label{fig:objects_bowl}
	\end{subfigure}
	\caption{Recording setup.}
	\label{fig:objects}
\end{figure}

To test the performance of the selected models on human manipulation actions we recorded a dataset of simple manipulation actions using a multi camera recording setup (see Figure~\ref{fig:setupIRL}). Movements were recorded with five calibrated and synchronized cameras at 100\,Hz with a resolution of $1024\times768$ pixels. Movement trajectories were extracted as described in Section \ref{sec:3Dtracking} below.

\subsection{Manipulation actions}
\label{sec:actionselection}
We selected nine different manipulation actions from \cite{Worgotter_SimpleOntologyManipulation_2013} all of which included only one object, which was actively manipulated. All actions are listed with a short description in Table \ref{tab:moves} (there also is a supplementary video with demonstrations).

\begin{table*}[ht]
	\centering
    \caption{List of the selected movements and short descriptions.}
    \label{tab:moves}
	\renewcommand{\arraystretch}{1.2}
    \small
	\begin{tabular}{ll}
		Movement					& Description\\
		\hline
		\hline
		\textit{pick and place}			& Pick up the manipulation object and place it at the target position.\\
		\hline
		\textit{put on top} / \textit{take down}	& Put the manipulation object on top of the target object /\\&take it down on the table at the start position.\\
		\hline
		\textit{put inside} / \textit{take out}	& Put the manipulation object inside the bowl-like target object /\\&take it out and place it on the table at the start position.\\
		\hline
		\textit{hide} / \textit{uncover}			& Put the manipulation object over the smaller target object /\\&take if off and place it on the table at the start position.\\
		\hline
		\textit{push} / \textit{pull}				& Push the manipulation object along the table without lifting it to the target position /\\&pull it back to the start position.\\
	\end{tabular}
\end{table*}

We used five different objects in the experiments as shown in Figure \ref{fig:objects_bowl}. A red wooden cube of dimensions 8\,cm x 8\,cm x 6\,cm was used as the main manipulation object. It was open at the bottom and hollow inside, such that it could also be used to hide and uncover a smaller blue cube.

Most of the actions also included a second fixed object, called the ``target object''. We used three bowl-like target objects of different height for the \textit{put on top}/\textit{take down} and \textit{put inside}/\textit{take out} actions. These objects were turned around to give raised bases of different height for the \textit {put on top} and \textit{take down} actions.

\begin{figure}[ht]
	\centering
	\includegraphics[width=0.9\linewidth]{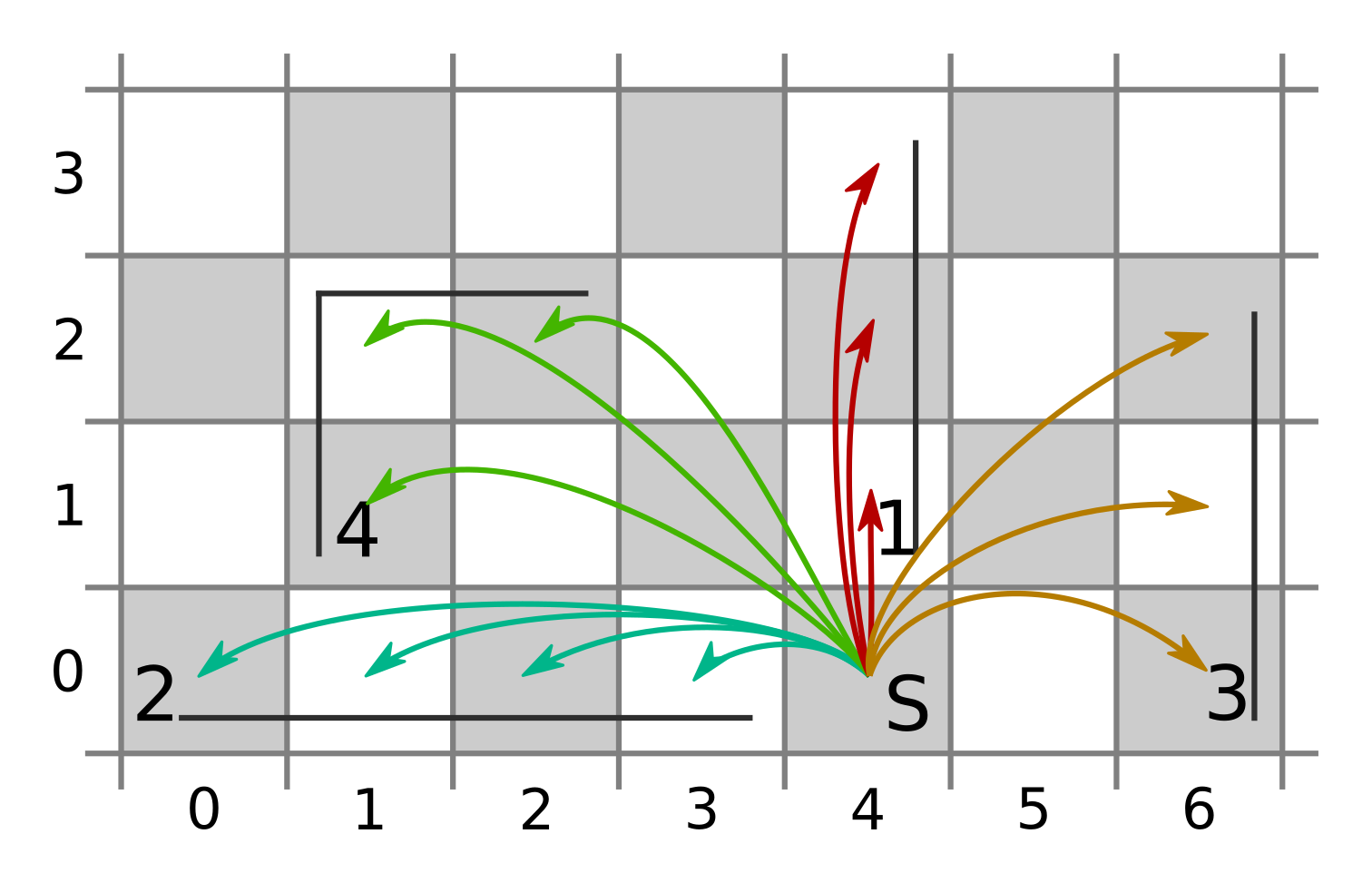}
	\caption{Schematic representation of the selected movement target positions on a grid with 10\,cm spacing (see also Figure \ref{fig:exampleview}). The start/end position is labeled S, the target positions are grouped into groups indicated by color and numbers.}
	\label{fig:placemat}
\end{figure}

To control the start and end positions of the movements, we restricted them to a square grid with 10\,cm grid spacing and chose the target positions to cover a range of different directions as shown by arrows in Figure \ref{fig:placemat}. The movements in group 1 are frontal, in group 2 lateral, and groups 3 and 4 contain moves in mixed directions.

We obtained data from 12 adult participants, 4 females and 8 males. Each participant performed nine actions as detailed in Table \ref{tab:moves} to the 13 different start-/end-points (see Figure \ref{fig:placemat}).
When using one of the bowl-like target objects, the moves to (3,0) and (4,1) were omitted due to space constraints, leaving only 11 start-/end-points. Each movement was repeated three times to accommodate for variability in the human demonstrations. Note that in some cases accidentally people performed two or four demonstrations instead of three. In total, our dataset consisted of 7,652 movements\footnote{The dataset is published on Zenodo:\\ \url{https://doi.org/10.5281/zenodo.7351664}}.

\subsection{Extraction of the 3D trajectories}
\label{sec:3Dtracking}

From the obtained movement recordings, we extracted the positions of the knuckles and wrist in the 2D images using \texttt{DeepLabCut} (DLC) \cite{Mathis_DeepLabCutmarkerlesspose_2018}, a neural network, which can be trained to track user-specified features in images and is publicly available as a python package \cite{Nath_UsingDeepLabCut3D_2019, _DeepLabCutDeepLabCut_}.
For the training of \texttt{DLC}'s neural network, we labeled 1000 frames out of all recordings.
They were chosen by \texttt{DeepLabCut} using its internal clustering algorithm for extraction of the most differing frames in the respective videos.
The error in feature tracking obtained from the test set is around three pixels.
This corresponds to up to 5\,mm distance, depending on the position of the hand in the camera frame.
An error of 5\,mm corresponds roughly to the size of the knuckles, which are inherently not very strictly defined features.

\begin{figure}[ht!]
	\centering
    \captionsetup[subfigure]{oneside,margin={0.8cm,0cm}}
	\begin{subfigure}[t]{0.49\linewidth}
		\includegraphics[width=\textwidth]{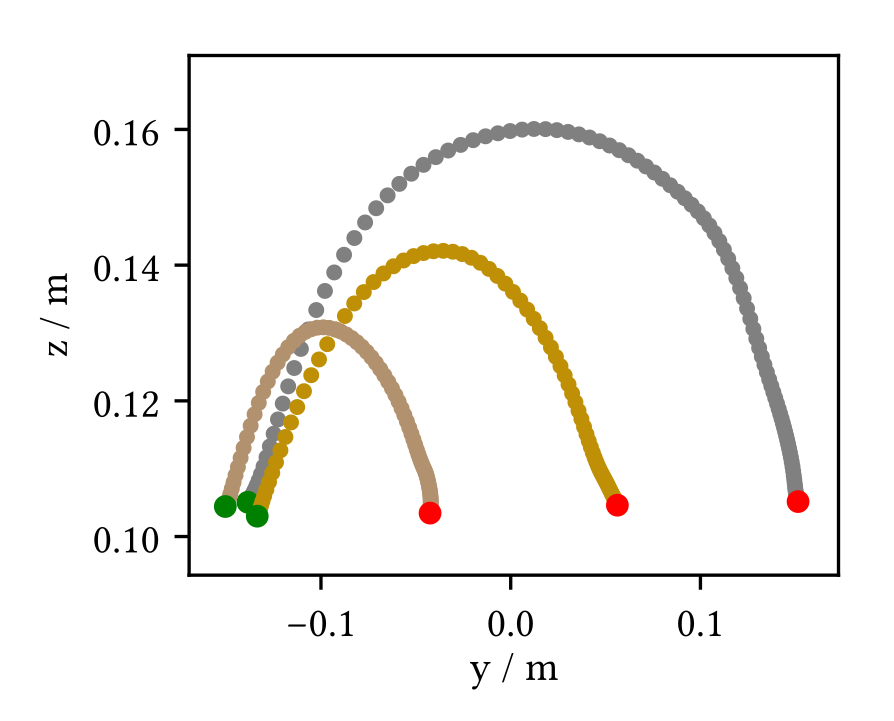}
		\subcaption{\normalfont \textit{Pick and place} to \\different target positions.}
		\label{fig:exampletrajectories_pap}
	\end{subfigure}
	\begin{subfigure}[t]{0.49\linewidth}
		\includegraphics[width=\textwidth]{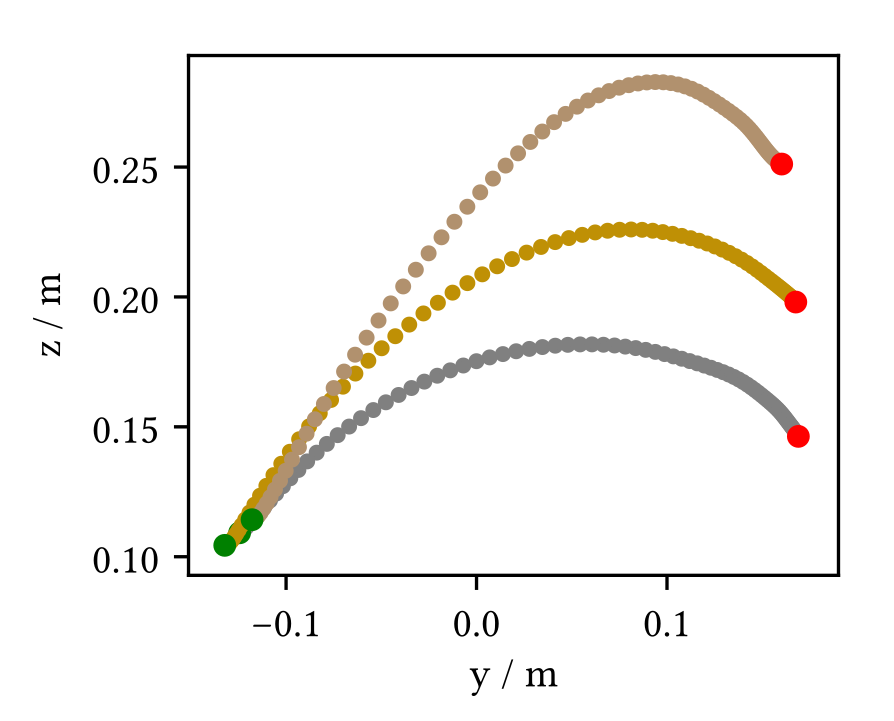}
		\subcaption{\normalfont \textit{Put on top} of objects \\with different heights.}
		\label{fig:exampletrajectories_pot}
	\end{subfigure}
	\begin{subfigure}[t]{0.49\linewidth}
		\includegraphics[width=\textwidth]{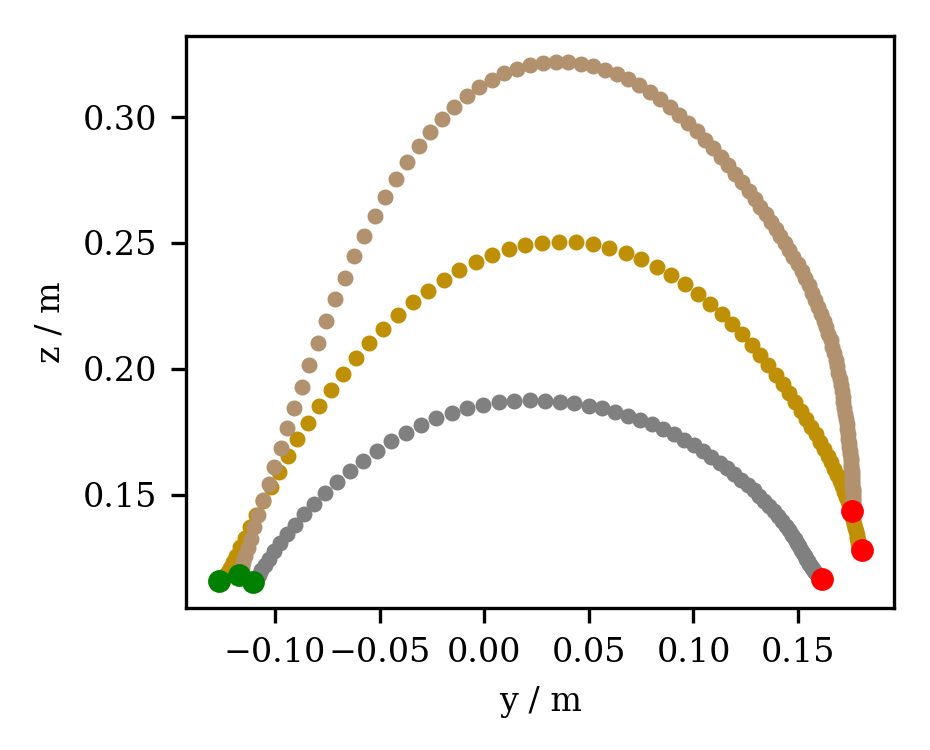}
		\subcaption{\normalfont \textit{Put inside}  objects \\with different heights.}
		\label{fig:exampletrajectories_pui}
	\end{subfigure}
	\begin{subfigure}[t]{0.49\linewidth}
		\includegraphics[width=\textwidth]{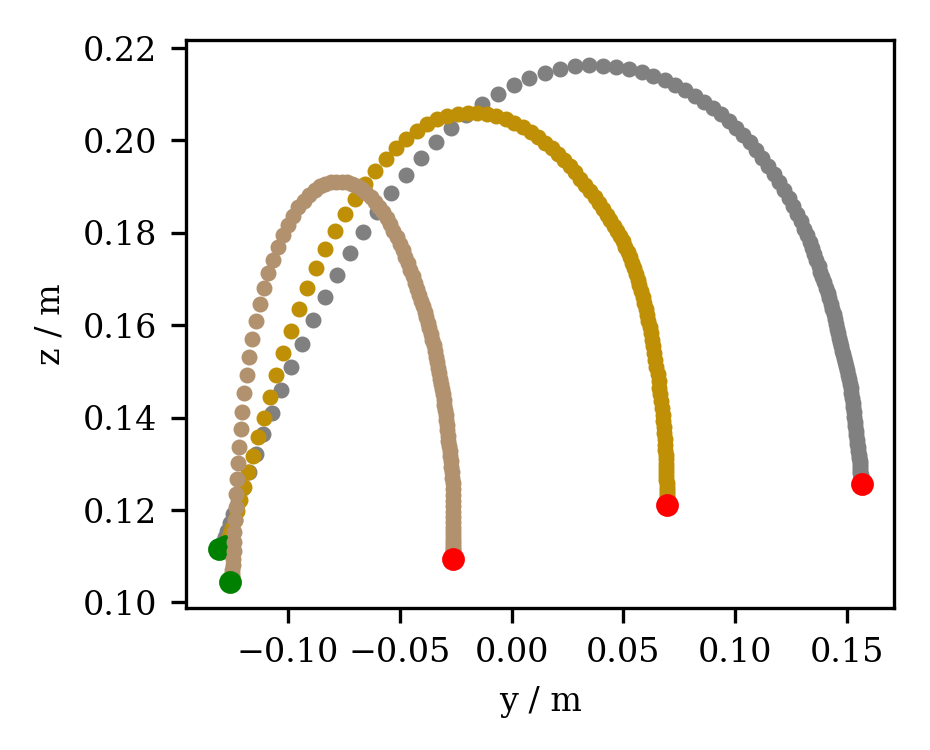}
		\subcaption{\normalfont \textit{Hide} to different \\target positions.}
		\label{fig:exampletrajectories_hid}
	\end{subfigure}
	\caption{Examples of trajectories from the movement dataset. Green and red dots denote start- and end-points, respectively.}
	\label{fig:exampletrajectories}
\end{figure}

After 2D tracking, we reconstructed the 3D positions of those features using \mbox{\texttt{anipose} \cite{Karashchuk_Aniposetoolkitrobust_2020}}, a publicly available 3D reconstruction library which is specifically built to work with \texttt{DeepLabCut}  \cite{_lambdaloopanipose_}.
We used its \textit{spatiotemporally constrained triangulation} method for the reconstruction of the 3D trajectories.
It introduces additional terms in the triangulation loss function that penalize behavior that is possible in general, but not in the special case of hands.
Tracked points may not accelerate arbitrarily fast (hands have mass) and their relative position cannot change arbitrarily, because hands have a given physical structure.
To obtain the final trajectories used in this work, we averaged the trajectories of the four knuckles and the wrist into one general hand position.
Some example motions are show in Figure \ref{fig:exampletrajectories}.

\section{Movement Encoding Frameworks}
The sections on DMPs, GMMs and SEDS were shortened to include less detail.
\subsection{Dynamical Movement Primitives}

\subsubsection{Model description}
The framework of Dynamical Movement Primitives (DMPs) \cite{Ijspeert_DynamicalMovementPrimitives_2013, Schaal_Dynamicssystemsvs_2007} is formalized by a system of second order differential equations.
They consist of a critically damped linear attractor system and a non-linear forcing term $f$ to make the attractor follow a desired trajectory.
The forcing term is learned from the presented data.
The DMP model is a 1D system, so three separate models need to be trained for 3D trajectories, one for each dimension.

Here we use the DMP version introduced by \cite{Kulvicius_JoiningMovementSequences_2012}.
It has been shown to have equivalent overall performance as compared to the original Ijspeert formulation \cite{Ijspeert_Movementimitationnonlinear_2002}, but better convergence at the end- and joining-point of the movement trajectory.
Since in this work we are not concerned with the analysis of joining of movement trajectories, we removed the parameter $v$ from the model (see Appendix \ref{app:DMPSigmoid}).
The forcing term $f$, that encodes the trajectory, is represented as a linear combination of $K$ Gaussian kernels with weights $w_i$.
For this work, we chose the kernels to be equally spaced in time.
Furthermore, we do not learn the individual standard deviations $\sigma_i$, but set them to scale with the number of kernels $K$ as 
\begin{align}
\sigma = \frac{\kappa}{K}, \label{eq:dmphyper}
\end{align} 
with some optimal $\kappa$ determined from data (see Appendix \ref{app:hyperparametersdmp}).
This means, the more kernels there are, the narrower they get, hence the sum of the standard deviations $\sigma_i$ remains fixed at $\kappa$.

The only individually learned parameters in this model are the weights $w_i$.
To learn the weights, locally weighted regression \cite{Ijspeert_DynamicalMovementPrimitives_2013} is used widely, but any other supervised learning rule can be used, too.
Here we used the simple $\delta$-learning rule as explained in \cite{Kulvicius_JoiningMovementSequences_2012}.

The number of free parameters in the model is just the number of weights, which is equal to the number of kernels $K$.
With one weight vector for each dimension $D$ we get $n_\text{DMP} = D K$ for the number of free parameters.
In our case of $D = 3$ this yields
\begin{align}
n_\text{DMP} = 3 K. \label{eq:freeparamdmp}
\end{align}

\subsubsection{Generalization using DMPs}
\label{sec:dmpgeneralizationmethod}

The simplest way to generalize to a new situation is to change the start $y_0$ or goal point $g$ of the DMP \cite{Ijspeert_DynamicalMovementPrimitives_2013} which also scales the trajectory accordingly based on the distance between the original and new star-/end-point.
However, this uniform scaling is not well suitable in all cases.
Consider a situation as given in \ref{fig:exampletrajectories_pot}~(b).
The \textit{put on top} trajectories all rise about 3\,cm above the end-point height before the object is set down, regardless of the actual end-point height.
If one would simply change the DMP goal point of the lowest trajectory to the highest end-point to achieve a generalization, the trajectory component for height would just scale accordingly (approximately by the factor of three, since the highest point is about three times higher).
This scaled trajectory would therefore rise around 9\,cm higher above the goal point before going down to reach the end-point, which would deviate by far (instead of the observed 3\,cm) from the human demonstration trajectory.

To achieve a generalization closer to the human trajectory, we instead take a weighted average of the DMP weight vectors $\bm{w}_n$ of nearby demonstrations.
This has been proposed before, but with weights only dependent on the distance to the demonstration targets \cite{Weitschat_Dynamicoptimalityrealtime_2013}.
We compare the performance of this method in Appendix \ref{app:DMPGeneralization}.
For our comparison, the weights $\alpha_n$, are obtained by the following process.
First, we shift every trajectory to start point $y_0 =0$.
Then, we express the new goal $g_\text{new}$ as a linear combination of the demonstration goals $g_n$ 
\begin{align}
    g_\text{new} &= \sum \alpha_n g_n
\end{align}
with averaging weights $\bm{\alpha} = (\alpha_1, ..., \alpha_n)$.
This equation is underdetermined, so we add the additional constraints
\begin{align}
    \sum \alpha_n &= 1 \text{ and } \vert \vert \bm{\alpha} \vert \vert \text{ minimal.}
\end{align}
Minimizing the norm of $\bm{\alpha}$ forces the demonstrations to contribute as equally as possible to the weighted sum.
This selects the one solution where all demonstrations contribute as much as individually possible to obtain the target trajectory (using the most of the available information), while still optimizing for giving more weight to the trajectories of highest relevance.
Else, as the system is heavily under-determined, it could be that only a few of the given demonstrations are considered at all.

Furthermore, we take an unweighted average over the dimensions that we do not change in the generalization.
The variation in the $g_n$ of those dimensions only comes from the demonstration variance and does not imply that the actual situation changed and some trajectories should contribute more than others.

Because of this mode of generalization, there is no fixed number of free parameters.
Instead it depends on the number of  demonstration encodings used, which each contribute $3K$ parameters according to Equation \eqref{eq:freeparamdmp}.
After averaging, the model again only contains $3K$ parameters, but cannot generalize to new situations on its own.

In addition to the main comparison between the three different models, we also performed a comparison between three different DMP generalization schemes: 1) change of the end-point, 2) weighted distance averaging, and 3) weighted goal averaging with constraints (as explained above). For the results of this comparison please see Appendix \ref{app:DMPGeneralization}.

\subsection{Task Parameterized Gaussian Mixture Models}

\subsubsection{Model description}
\label{sec:tpgmmgeneral}

Gaussian Mixture Models (GMMs) are  generally used for regression of arbitrary $D$-dimensional distributions.
A GMM performs regression on a general distribution $\bm{\xi}_t$ (with $t = 1, ..., N$ datapoints) by approximating it with a linear combination of $i = 1, ..., K$ multivariate Gaussian kernels $\mathcal{N}_i(\bm{\mu}_i, \bm{\Sigma}_i)$.
Here, $\bm{\mu}_i$ is the mean of a Gaussian (vector of length D), and $\bm{\Sigma}_i$ its variance ($D \times D$ matrix).
The linear combination coefficients are called $\pi_i$ and fulfill $\sum_i \pi_i = 1$.

There are multiple ways to encode trajectories using GMMs \cite{Calinon_tutorialtaskparameterizedmovement_2016, Calinon_improvingextrapolationcapability_03, Khansari-Zadeh_BMiterativealgorithm_2010, Khansari-Zadeh_LearningStableNonlinear_2011}.
In this work, we used the method called time-based Gaussian Mixture Regression (tbGMR, \cite{Calinon_tutorialtaskparameterizedmovement_2016}), which will be described in Section \ref{sec:tbGMR}.
This approach in itself can only encode trajectories and is unable to generalize from demonstrations to new situations.
To allow for that, a model called Task Parameterized GMM (TP-GMM) \cite{Calinon_tutorialtaskparameterizedmovement_2016} is used.
This model fits multiple models with the tbGMR approach simultaneously and combines them to a single model.
Note that the TP-GMM approach is also able to support different kinds of GMM encodings.
For our analysis, we used a \texttt{Matlab} / 
\texttt{Gnu Octave} implementation of the model provided by \cite{Calinon_tutorialtaskparameterizedmovement_2016}.

\subsubsection{Time based GMR}
\label{sec:tbGMR}

For time-based GMR, the joint distribution $\mathcal{P}(\bm{\xi}^I_t, \bm{\xi}^O_t)$ of time $\bm{\xi}^I_t$ (superscript $I$ for input) and the spatial coordinates $\bm{\xi}^O_t$ (superscript $O$ for output) at that time is encoded.
Note that, in contrast to DMPs, time is treated as an additional input dimension to the model, not as an independent variable.

To reconstruct the encoded trajectory, the conditional probability of having spatial coordinates $\bm{\xi}_t^O$ at a given time $\bm{\xi}_t^I$, $\mathcal{P}(\bm{\xi}^O_t \vert \bm{\xi}^I_t)$, is used.
Mathematical details on how to obtain it from  $\mathcal{P}(\bm{\xi}^I_t, \bm{\xi}^O_t)$ are given in \cite{Calinon_tutorialtaskparameterizedmovement_2016} in Section 5.1.
With some simplification, this distribution is again a multivariate Gaussian for each $\bm{\xi}_t^I$. Thus, we can take the centers $\hat{\bm{\mu}}_t^O$ as the predicted trajectory of the model.
Furthermore, the values of $\hat{\bm{\Sigma}}\strut^O_t$ contain additional information about the uncertainties in the model, as they define uncertainty-ellipsoids around the predicted positions.
Such information can, for example, be used in robotic applications, to indicate how strongly a motion controller should enforce following the predicted trajectory of the $\hat{\bm{\mu}}_t^O$.

To avoid overfitting, the model has to be regularized.
The implementation we used in this work realizes this by adding a small constant $\epsilon$ to the main diagonal of the covariance matrices after each M-step in the fitting algorithm.
This reduces their capability to get too restrictive in individual dimensions, smoothing out the obtained trajectories.

The parameters used by tbGMR are
\begin{align}
\left\{\pi_i, \bm{\mu}_i, \bm{\Sigma}_i \right\}_{i=1}^K. \label{eq:paramtbGMR}
\end{align}
The number of parameters depends on the number of Gaussians $K$ and the data dimension $D$.
Since the $\bm{\Sigma}_i$-matrices are symmetric, they have $(D^2-D)/2 + D$ free parameters each (the upper right triangle including the diagonal).
The $\bm{\mu}_i$ each contain $D$ free parameters.
From the $K$ linear combination coefficients $\pi_i$ only $K-1$ are free, because of the normalization condition $\sum_i \pi_i = 1$.
Thus, we have 
\begin{align}
n_\text{GMM} = K \left(\frac{D^2 - D}{2} + 2D\right) + K - 1 \label{eq:paramtbGMR_amount}
\end{align}
free parameters in total.
In the case of 3D-trajectories, this model is 4-di\-men\-sion\-al (one additional time dimension).
Therefore, setting $D=4$ into Equation \eqref{eq:paramtbGMR_amount}, we get
\begin{align}
n_\text{tbGMR} &= 15 K - 1
\end{align}
for the number of free parameters in the model. 

\subsubsection{Task parameterized GMM}
\label{sec:tpgmm}

A TP-GMM considers the data $\bm{\xi}$ in different reference frames $j = 1, ..., P$. They are obtained by affine linear transformations with some rotation matrices $\bm{A}_j$ and translation vectors $\bm{b}_j$.
Those are called the ``Task Parameters'', because they parameterize the relation of the different reference frames for each demonstration (task).
GMMs in all reference frames are fitted to the data using the Expectation-Ma\-xi\-mi\-za\-tion-Algorithm (EM-Algorithm, \cite{Dempster_MaximumLikelihoodIncomplete_1977}) in a modified version to fit all reference frames simultaneously (see Appendix A in \cite{Calinon_tutorialtaskparameterizedmovement_2016}).

The models in the different reference frames $P$ can be combined to one model from which a trajectory can be retrieved by means of the chosen encoding type (for more details see \cite{Calinon_tutorialtaskparameterizedmovement_2016}).
The ability to generalize to new situations is introduced by combining the models of the different reference frames with different task parameters than the ones used during learning.
Note that the individual GMMs in the two different reference frames encode something conceptually different from a standard GMM with one reference frame. They cannot be used individually to retrieve the encoded trajectories, since they represent a combination of all learned trajectories as seen from one reference frame only. 

It is possible to take different kinds of GMMs for encoding and reconstruction and enable generalization via the TP-GMM approach.
Here we used time based Gaussian Mixture Regression as explained in Section \ref{sec:tbGMR} to encode trajectories.
The usage of $P$ different reference frames leads to $P$ sets of $\bm{\mu}_i^{(j)}, \bm{\Sigma}_i^{(j)}$ in the model instead of one.
Thus, we have 
\begin{align}
n_\text{TP-GMM} = K P \left(\frac{D^2 - D}{2} + 2D\right) + K - 1 \label{eq:paramTPGMM_amount}
\end{align}
free parameters in total.
This only differs from Equation \eqref{eq:paramtbGMR_amount} by the factor of $P$ in the first term reflecting the additional sets of $\bm{\mu}_i^{(j)}, \bm{\Sigma}_i^{(j)}$.
In this work we will use $P=2$, so putting in $D=4$ from out tbGMR approach and $P=2$ we get
\begin{align}
n_\text{TP-GMM} &= 29 K - 1
\end{align}
free parameters for a TP-GMM.

\subsubsection{Generalization using TP-GMM}
\label{sec:tbgmr_generalization}

In contrast to the DMP framework, where generalizations can be obtained by combining the weights of existing encodings for single trajectories, the TP-GMM approach operates on new encodings of multiple demonstration trajectories.

To generalize from a set of demonstrations to a new situation, we encode a TP-GMM with two reference frames $P=2$.
We chose the Task Parameters such that they rotate and translate all demonstration trajectories into two reference frames defined as follows:
In the first frame, identified by $j=1$, all trajectories begin at $\bm{0}$.
Furthermore, they are rotated such that their end-point lies in the \textit{yz}-plane. 
In the second frame, denoted by $j=2$, the roles of start- and end-point are switched.
This yields $\bm{A}_j$ and $\bm{b}^O$ as needed by TP-GMM.

To generalize to a new trajectory, the recombination to one GMM is done with the Task Parameters of the target trajectory. 
These are obtained in exactly the same way as for the demonstration trajectories.
The resulting GMM encodes a prediction for a trajectory with the new Task Parameters, which can be extracted by means of the  encoding method used.

\subsection{Stable Estimator of Dynamical Systems (SEDS)}
\label{sec:seds}

\subsubsection{Model description}
\label{sec:seds_model}
The SEDS model \cite{Khansari-Zadeh_LearningStableNonlinear_2011}, like the DMP approach, considers the trajectory to be the time evolution of a nonlinear dynamical system.
In the case of SEDS it takes the simple form
\begin{align}
    \dot{\bm{\xi}} = \bm{f}(\bm{\xi})
\end{align}
where $\bm{\xi}$ is the position in state space and $\bm{f}$ some multi-dimensional nonlinear function.
Trajectories cannot cross in state space, so care needs to be taken when choosing the state space variable $\bm{\xi}$.
In this work we used position and velocity.
The function $\bm{f}$ gets approximated by a GMM analogous to the time based GMR method from Section \ref{sec:tbGMR}.
The joint distribution $\mathcal{P}(\bm{\xi},\dot{\bm{\xi}})$ is encoded with a GMM.
For given $\bm{\xi}$, $\dot{\bm{\xi}}$ is then extracted as the maximum of the conditional probability $\mathcal{P}(\dot{\bm{\xi}}\vert\bm{\xi})$.

In general, there exist several methods to estimate the parameters of the GMM (see Figure 3 in \cite{Khansari-Zadeh_LearningStableNonlinear_2011}), but they all have problems with stability.
It is, thus, not guaranteed that a reconstructed trajectory will converge to the goal position in case of GMM.

The core of the SEDS approach is its way of estimating $\bm{f}$.
With SEDS, it is guaranteed that there is only one global attractor in the dynamic system, that is asymptotically stable, so trajectories from every point will converge to it.
This property is achieved by reformulating the estimation of the GMMs parameters as a constrained nonlinear optimization problem optimizing model accuracy under the constraint of global asymptotic stability.
The exact formulation of the optimization problem is given in Section V in \cite{Khansari-Zadeh_LearningStableNonlinear_2011}.
As shown in Figure 3 in \cite{Khansari-Zadeh_LearningStableNonlinear_2011}, the SEDS optimization leads to a smooth phase space with no secondary attractors.
There, two different measures for model accuracy were proposed, the likelihood of the model and the mean squared error (MSE) of its prediction to the demonstration. In this work, we used the MSE.
\cite{Khansari-Zadeh_LearningStableNonlinear_2011} also provides an implementation of the algorithms, which we used for our analysis.

While the the resulting model is guaranteed to converge to the goal state, the nonlinear optimization to fit the model is not guaranteed to succeed.
Our strategy of dealing with this problem is described in Appendix \ref{app:hyperparametersseds}.

The number of free parameters in a SEDS model optimized with the MSE-loss is $K(1 + \frac{2}{3} D(D + 1))$ \cite{Khansari-Zadeh_LearningStableNonlinear_2011}.
For our state space dimensionality of $D=6$ (position and velocity) this yields
\begin{align}
    n_\text{SEDS} = 29 K.
\end{align}

\subsubsection{Generalization using SEDS}
Since SEDS optimization yields a globally stable attractor, one can simply change the position of the starting point of the integration to obtain a new trajectory that will converge to the target.
While the model is designed to yield  trajectories similar to the demonstrations in a vicinity around them, it is unclear whether its predictions will be accurate for generalization targets farther away.
Therefore, the similarity between generalization and true human demonstrations will depend on the nature of the attractor landscape produced by the optimization.
Furthermore, the duration of the movement cannot be set directly, but also depends on the properties of the attractor.
In a first step we just integrate the dynamic system for as many timestamps as there are in the target trajectory (like we do with the other models).
Additionally we also integrate the model for as many timesteps as needed to arrive within 5\,mm of the target position, this distance corresponds to our estimated motion tracking accuracy in the dataset.

\subsection{Probabilistic Movement Primitives}
\label{sec:promp_model}

\subsubsection{Model description}
Probabilistic Movement Primitives (ProMPs) \cite{Paraschos_ProbabilisticMovementPrimitives_2013, Paraschos_probabilisticapproachrobot_2015, Paraschos_Usingprobabilisticmovement_2018} are formulated as a Hierarchical Bayesian Model (HBM).
A trajectory is modeled as a superposition of Gaussian basis functions with weights $\bm{w}$.
In contrast to the general GMM used in tbGMR (see Section \ref{sec:tbGMR}), the kernel shapes are set (as hyperparameter), not fit.
To encode the variance from multiple demonstrations, a distribution $p(\bm{w};\bm{\theta})$ over the weights is introduced.
By marginalizing $\bm{w}$ and assuming a Gaussian distribution of $p(\bm{w};\bm{\theta})$ we arrive at the distribution $p(\bm{y}_t\vert\bm{\theta})$ as defined by Equation (4) in \cite{Paraschos_ProbabilisticMovementPrimitives_2013}.
The parameters $\bm{\theta} = \{\bm{\mu}_w, \bm{\Sigma}_w\}$ are fit by a specialized version of the EM-Algorithm for HBMs \cite{Lazaric_Bayesianmultitaskreinforcement_2010}.
We use the implementation provided by the \texttt{movement\_primitives} python library \cite{DFKIGmbHRoboticsInnovationCenter_dfkiricmovement_primitives_2021}.

As with the tbGMR approach, one can not only determine the mean positions for each time point, but also variance around them, providing additional information, for example for improving the controller used to follow the trajectory.
Additionally, the model also encodes full crosss-covariances between all the dimensions.
The HBM nature of the encoding allows for changes to the distributions to be done by conditioning on new observations and propagating the results through to $\bm{\theta}$ by Bayes rule.
This way, via points or even new start/end positions can be introduced.

Regularizing via the same method as for tbGMR proved to be ineffective, so we regularized the model by setting the widths of the basis functions, removing the tendency to oscillate like the tbGMR.

The parameters of a ProMP are the $\bm{\mu}_w$ and $\bm{\Sigma}_w$.
Given $K$ basis functions, we have $DK$ free parameters from the means and $((DK)^2-DK)/2+DK$ from the covariances (symmetric $DK\times{}DK$ matrix). With $D=3$ this yields
\begin{align}
    n_\text{ProMP} = \frac{9}{2}(K^2+K).
\end{align}
For the reconstruction case, where only one demonstration is presented and no variance is encoded, the covariance matrix is only very sparsely populated and the number of parameters decreases to ${n_\text{ProMP}}_{rec} = \frac{1}{2}(K^2+5K)$.

\subsubsection{Generalization using ProMPs}
\label{sec:promp_generalization}

The ProMP framework provides multiple ways of generalization.
Different ProMPs can be joined or co-activated, to form new trajectories.
In our case, generalization was achieved by conditioning the ProMPs on new start positions as explained below.

Similar to the TP-GMM method, we learn a ProMP from multiple demonstrations, but not only of the same motion, but of different ones.
We shift all trajectories such that their end points are at $\bm{0}$.
The resulting model encodes the positions of trajectories and their variance at once.
Thus, like with the individual TP-GMM reference frames, evaluating the model would produce some ``mean of all demonstrations'' representation, not any single trajectory.
Conditioning on the start point of the generalization target trajectory fixes the model to its prediction of that trajectory.

\subsection{Optimal Control Primitives}

\subsubsection{Model description}
Optimal Control Primitives (OCPs) \cite{Herzog_Generationmovementsboundary_2017} consist of two main parts, a trajectory representation and a controller derived from optimal control theory.

The trajectory representation could theoretically just be the trajectory itself, but to achieve some sort of encoding and compressed representation, the trajectory is expanded into a series of Chebyshev polynomials up to a given order.
So, like with DMPs, the shape of the trajectory is encoded as a set of weights, just not for Gaussian kernels distributed through a phase variable (time), but as the summation coefficients of a series expansion.

The controller is used to follow the trajectory with perturbation resistance and enable adaptation to different scenarios.
It is a linear-quadratic-regulator (LQR) derived from optimal control theory.
The derivations and underlying models are given in \cite{Herzog_Generationmovementsboundary_2017}.
It can be tuned with three parameters.
$R$ decides if to follow the reference trajectory more strictly or loosely, with high $R$ meaning low control signals. Choosing $R$ too high will result in the controller not properly following the trajectory.
The other two parameters $q1$ and $q2$ weigh the controllers reaction to spatial accuracy or correct velocity.
$q1 > q2$ means more importance is placed on following the position profile than the velocity profile and vice-versa.

The number of free parameters in the model is equal to the number of weights for the trajectory representation.
The Chebyshev polynomial series starts with order 0, so a representation of order $o$ has $o+1$ weights.
Therefore, we chose the convention that an OCP with $K=n$ uses a Chebychev polynomial series up to order $n-1$.
This way, we have
\begin{align}
    n_\text{OCP} = 3K
\end{align}

\subsubsection{Generalization using OCPs}
Generalization using OCPs can be achieved by just changing the start and end points of the trajectories and letting the controller adapt to it.
However, OCPs have no scaling properties like DMPs, so the resulting trajectories will be qualitatively different. For further analysis of this refer to Appendix \ref{app:DMPGeneralization}.

Still, the same considerations regarding using multiple models for generalization apply, as with the DMPs in Section \ref{sec:dmpgeneralizationmethod}.
So, to enable the OCPs to use all available demonstrations, we used the same generalization method as described for DMPs in Section \ref{sec:dmpgeneralizationmethod} and computed weighted averages of weights of the Chebyshev series.

As a result, there is again no fixed number of free parameters.
Instead it depends on the number of demonstration used, but after averaging, the model again only contains $3K$ parameters.

\section{Comparison of Movement Frameworks}

\subsection{Movement Reconstruction}

\subsubsection{Evaluation procedure}

To test the reconstruction capabilities of the different models, we encoded every single trajectory from our dataset with every model, varying the number of kernels used from $K=3$ to $K=11$ for DMP, tbGMR, ProMP and OCP and from $K=3$ to $K=7$ for SEDS.
In case of SEDS, we did not use higher $K$ values as for the other models, since the encoding times due to the optimization process became unreasonably long.
The respective hyperparameters of the models were chosen optimally, detailed information on this is given in Section \ref{sec:hypertune_rec} and Appendix \ref{app:hyperparameters}.
We also investigated the influence of sub-optimally chosen hyperparameters on the reconstructions, because sensible output for a wider region of hyperparameters means better usability of the model.

\subsubsection{Performance measures}

To assess the error in the reconstructions, we computed the root mean squared distance $d$ between the reconstruction $r$ and the original trajectory $o$ for every reconstruction as

\begin{align}
    \label{eq:traj_deviation}
    d = \sqrt{\frac{1}{N} \sum \limits_i^N ( r_i - o_i )^2}.
\end{align}

The resulting distributions of errors are not normally distributed, but skewed and contain a tail of more inaccurate reconstructions.
Therefore we use medians instead of means to represent the average error in the reconstructions of the different models.

\subsection{Movement Generalization}

\subsubsection{Evaluation procedure}
\label{sec:generalization_procedures}
To test the generalization capabilities of the models, we presented specific subsets of the dataset (training set) as demonstrations for encoding and then used the generalization methods described above to generate trajectories for new situations obtained from remaining trajectories in the dataset (test set).
This way we can directly compare the model predictions to the trajectories of human demonstrations.

We tested two different generalization scenarios with different selection of demonstration trajectories.
In the first case we only used two trajectories of one action type to generate a new trajectory.
This represents the case where only few demonstrations are available. For this, we selected triplets of start-/end-points that were orthogonally connected (see Figure \ref{fig:placemat}).
We then interpolated from the outer demonstrations to the inner one and extrapolated from the inner and one of the outer demonstrations to the respective other outer demonstration.
For example, using the indexing from Figure \ref{fig:placemat}, we extrapolated from moves to positions (1,0) and (1,1) to a move to (1,2).
Because we have three repetitions of each action type, this means the models get 6 demonstration trajectories (two sets of three instances of the same class) to generate the new target trajectory.

In the second case we used all available demonstrations for a specific action type, but excluded the generalization targets and everything that ends closer than 3\,cm to the generalization target's end point.
For the TP-GMM-model we further excluded all demonstrations that were closer than 3\,cm to the target position in the TP-GMM reference frame, otherwise, trajectories for positions with the same length over ground (for example (2,2) and (6,2), see Figure \ref{fig:placemat}) would have provided direct demonstrations for the generalization target.

In both cases we did not use demonstrations from multiple participants in one model, every generalization task was done for each participant individually.

Again, model hyperparameters were chosen optimally, as described in Section \ref{sec:hypertune_gen} and Appendix \ref{app:hyperparameters}.

\subsection{Performance measures}
\label{sec:performance_measures}

As in the case of trajectory reconstruction, the main measure to assess the generalization quality is the deviation of the generalized trajectory from the human target demonstration.
We use the same measure as before, given in Equation~(\ref{eq:traj_deviation}).
To gain a more complete picture of the model performance, we used two additional measures for further comparison.

We compared the deviation of the generalization from the true demonstration with the general variance in the dataset itself.
This helps to qualify meaning of the deviation between generalized and ground truth (human) trajectory, since identical trajectories (as in reconstruction) are not expected.
It is possible to do this, because we have three repetitions of every single action.
So to estimate the variance in the human execution of the actions, we measured the mean squared distances $d_h$ between the individual human repetitions $h^{(j)}$ as

\begin{align}
d_h^{(j,k)} = \sqrt{\frac{1}{N} \sum \limits_i^N ( h^{(j)}_i - h^{(k)}_i )^2}.
\end{align}

To compare the trajectories of different lengths, we down-\-sam\-pled the longer trajectory to match the shorter trajectory.

Note that $d$ and $d_h$ are conceptually different measures.
$d$ measures error between a human trajectory and the corresponding generalized trajectory, whereas $d_h$ measures variance in the human repetitions.  
The end-points of the human trajectories were not given, because humans grasped the object slightly differently in different repetitions, resulting in differences between the repetitions.
The model generalizations, however, are given exact start- and end-points.
Therefore, the inter human variance for repetitions of a demonstration is conceptually different and expected to be higher than the distance between a generalized target trajectory and the corresponding human demonstration.
We assume that the obtained generalization accuracy is reasonably acceptable if it does not exceed the variance of the human demonstrations.

The second measure we used for additional quantification was the end-point deviation $d_e$, defined as
\begin{align}
    d_e = \lVert r_N - o_N \rVert
\end{align}
with the predicted trajectory $r$ and the original trajectory $o$, both of length $N$.
Here we assessed how close the end-points of the generalization come to the targets set by the human demonstration, regardless of the trajectory shape.
This is to test the end-point convergence of the two GMM models.

\begin{figure*}[ht!]
\centering
\begin{subfigure}{0.3\textwidth}
    \includegraphics[width=\textwidth]{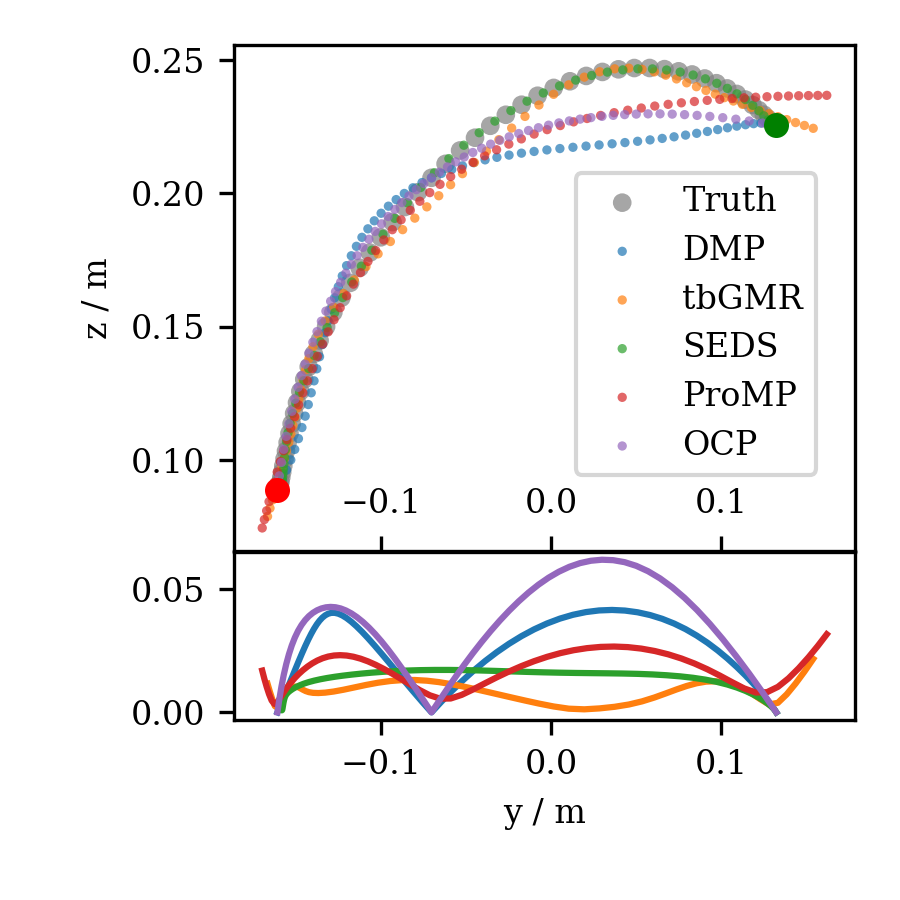}
\end{subfigure}
\hfill
\begin{subfigure}{0.3\textwidth}
    \includegraphics[width=\textwidth]{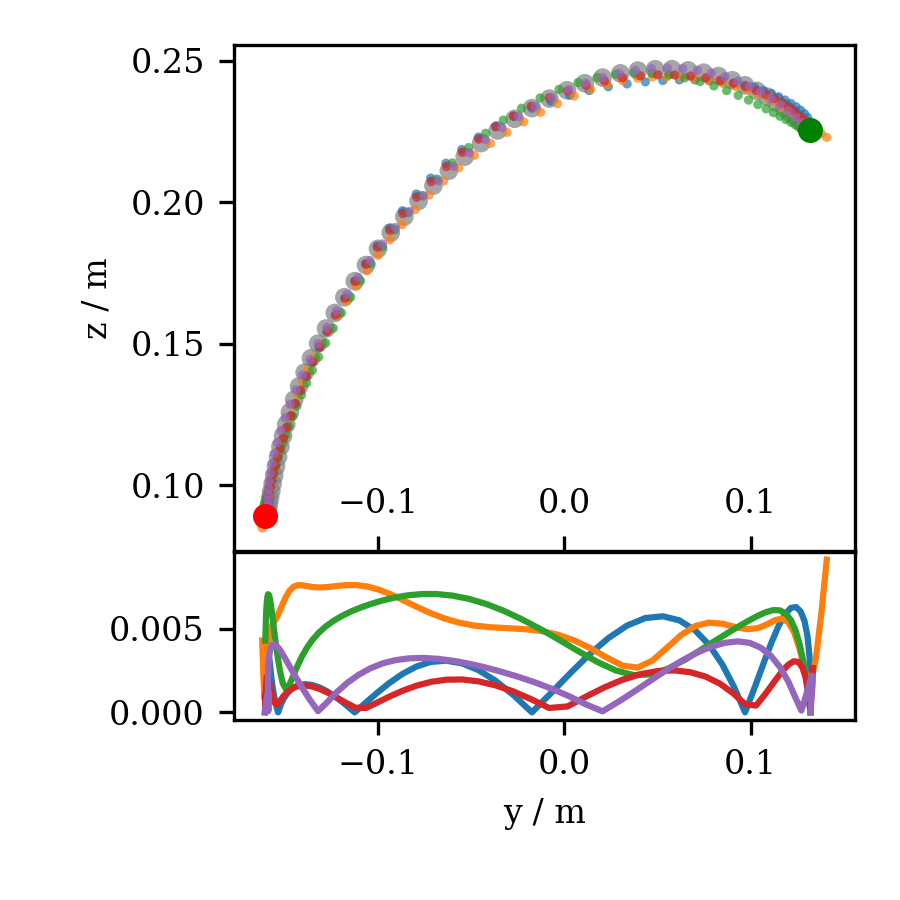}
\end{subfigure}
\hfill
\begin{subfigure}{0.3\textwidth}
    \includegraphics[width=\textwidth]{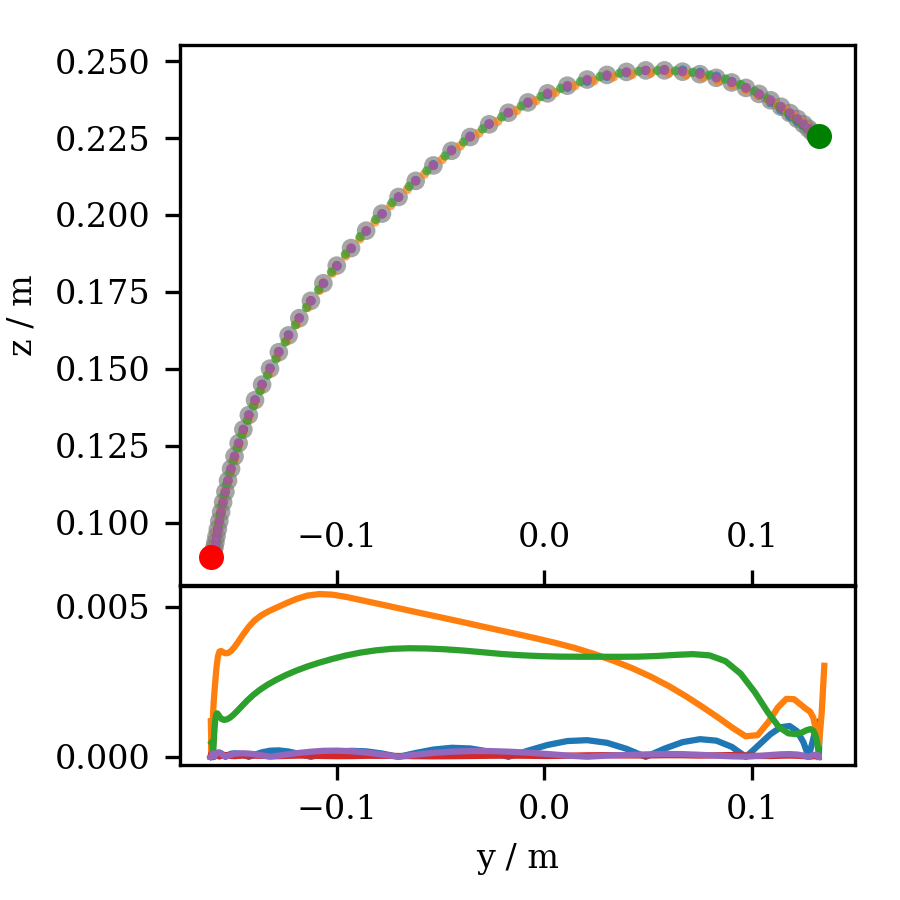}
\end{subfigure}
\\
\begin{subfigure}{0.3\textwidth}
    \includegraphics[width=\textwidth]{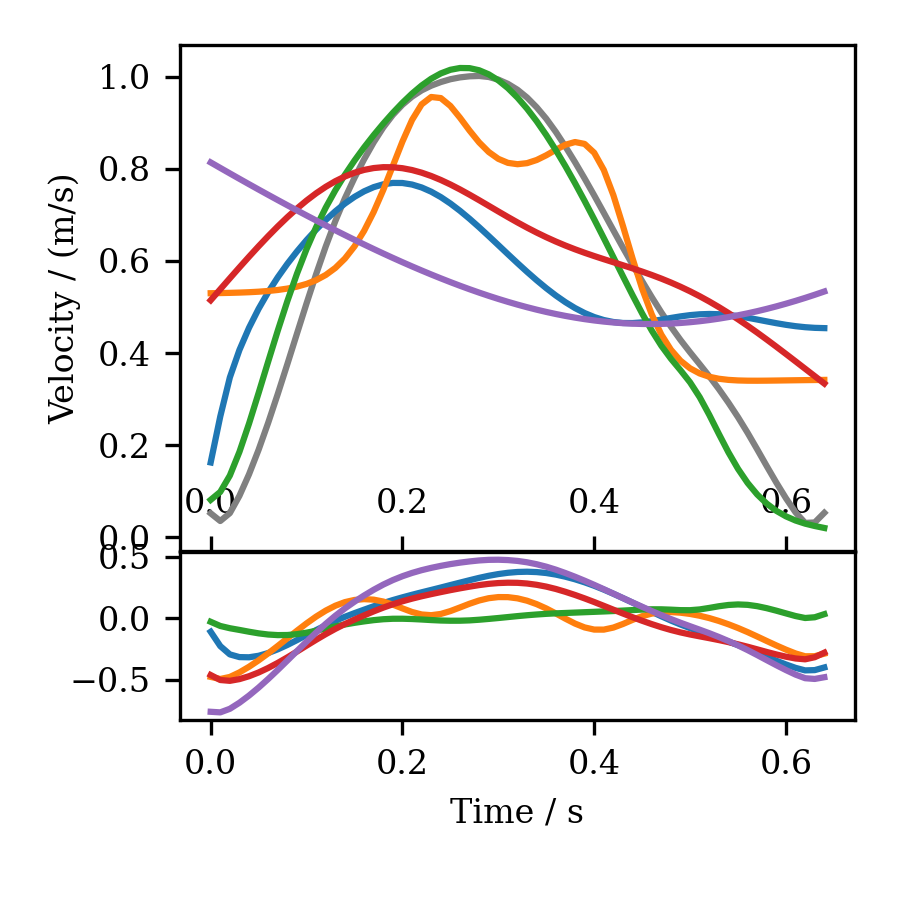}
    \caption{3 kernels.}
    \label{fig:recexamples_few}
\end{subfigure}
\hfill
\begin{subfigure}{0.3\textwidth}
    \includegraphics[width=\textwidth]{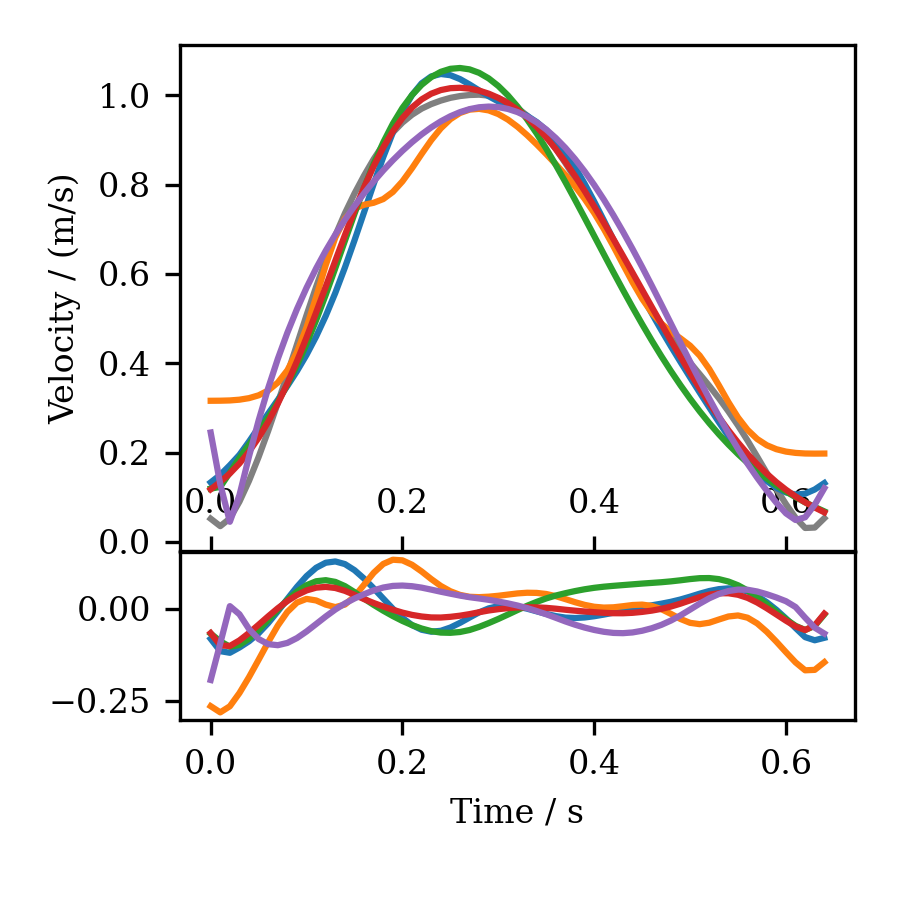}
    \caption{6 kernels.}
    \label{fig:recexamples_med}
\end{subfigure}
\hfill
\begin{subfigure}{0.3\textwidth}
    \includegraphics[width=\textwidth]{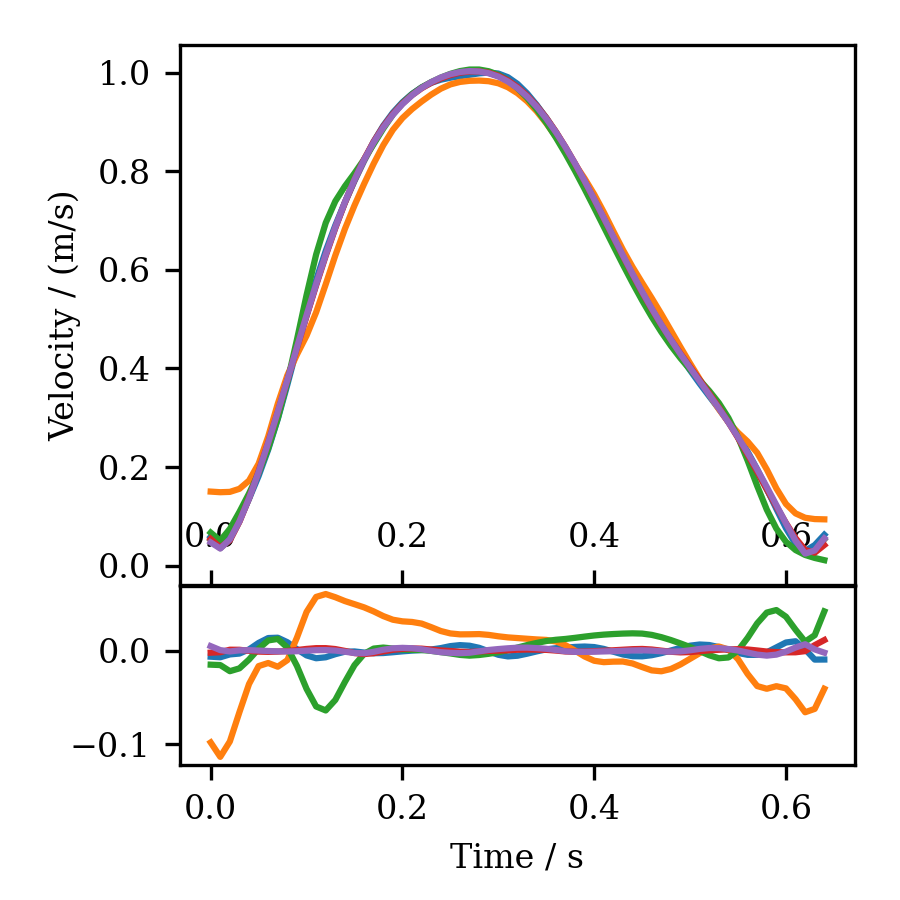}
    \caption{11 kernels.}
    \label{fig:recexamples_hig}
\end{subfigure}
\caption{Example of human trajectories of a \textit{take down} action and reconstructions  for different numbers of kernels: 3 (a), 6 (b) and 11 (c).
Position and velocity profiles are shown in the top and bottom rows, respectively, whereas the plots at the bottom of each panel show the deviation from human the trajectory. Green and red dots denote start- and end-points, respectively.
}
\label{fig:recexamples}
\end{figure*}

Since the SEDS model does not always converge to the target position \textit{on time} (see Section \ref{sec:modeldifferences}), we evaluated the generalization performance of SEDS for two cases: 
1) under the condition with fixed trajectory duration (labelled SEDS in figures) as for the other models.
Here, the fixed trajectory duration corresponds to the duration of the human demonstration.
2) under the relaxed condition on trajectory duration (labelled SEDS*). 
For this, we first did an estimation of acceptable time convergence.
We measured how much longer the longest of the three repetitions of a human movement takes compared to the shortest one.
We found that 95\,\% of all longest human demonstrations completed in less than 38\,\% more time than the shortest demonstration for the same task.
Therefore, we counted a SEDS generalization successful under the relaxed time condition if it completed within 72.5\,\% and 138\,\% of the duration of the human demonstration and computed the generalization error only among those successful generalizations.

\section{Results}

\subsection{Movement Encoding and Reconstruction}

\subsubsection{Hyperparameter tuning}
\label{sec:hypertune_rec}

Choosing sub-optimal hyperparameters has different effects, depending on the type of model.
The stability of the SEDS model is not considered here but in Section \ref{sec:sedsconvergence}, because we don't have direct influence on the kernel width in that model.

DMP, tbGMR and ProMP will produce oscillations in the velocity profile, if the kernel width is too narrow in case of DMPs and ProMPs or the regularization term is too small in case of tbGMR.
This is not the case for the human demonstrations and should therefore be avoided.
Choosing sub-optimal width for the DMP and ProMP kernels increases the reconstruction error (see Appendix \ref{app:hyperparameters}).
For tbGMR, however, lowering the regularization, and therefore introducing oscillations in the velocity profile, reduces the reconstruction error of the position profile.
This makes choosing hyperparameters more difficult, as the well defined reconstruction error cannot be the only criterion (see Appendix \ref{app:hyperparameterstbgmr}).
Therefore it is harder to determine the optimal regularization for accurate but still human-like trajectories with the tbGMR model.

\subsubsection{Reconstruction accuracy}
\label{sec:rec_accuracy}

Figure \ref{fig:recexamples} shows some examples of trajectory reconstructions using different numbers of kernels that are indicative of some general behaviour of the models.

With only three kernels, all models but SEDS struggle with accurate trajectory representation.
Although OCP and DMP have larger overall error than the other models, they have zero error at the start- and end-point, which is a direct outcome of their respective learning mechanisms.
The tbGMR and ProMP models even miss the correct start- and end-point completely.
This behaviour is inherent and can't be avoided.

When increasing the number of kernels, the reconstructions become qualitatively identical to the demonstrations with errors dropping below the 5\,mm of tracking error in all cases.
The main source of error for tbGMR and SEDS at high $K$ is not the trajectory shape, but rather deviations in velocity.
In contrast to all other models, tbGMR is not able to start the trajectory with near zero velocity, resulting in it being ahead of the target.

\begin{figure}
    \centering
    \includegraphics[width=\linewidth]{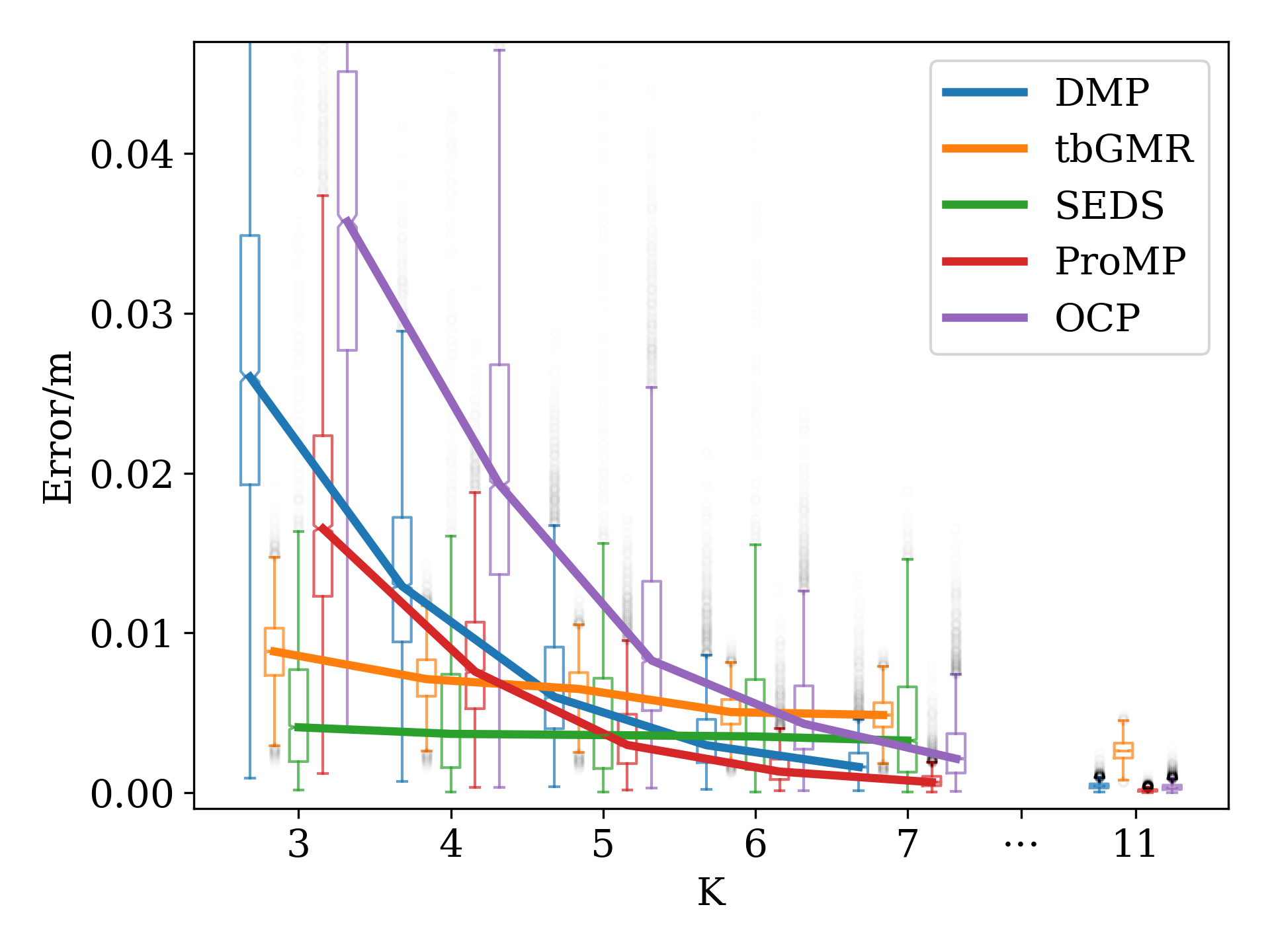}
    \caption{Comparison of movement encoding frameworks on the movement reconstruction task. Median reconstruction error vs. number of kernels is shown for each model.
    Failed SEDS encodings are not included in this statistics (see Section \ref{sec:sedsconvergence}).
    Note, there is no result for SEDS for $K=11$ because the encoding time for high $K$ values using the repeated optimization method (see Section \ref{sec:seds_model}) is impractically long.}
    \label{fig:recerror}
\end{figure}

The median reconstruction error obtained from all reconstructions vs. the number of used kernels is shown in \mbox{Figure \ref{fig:recerror}}.

We can see that the median errors of all models fall below our 5\,mm tracking error estimate, if a sufficient number of kernels is used.
However, there are qualitative differences among the five models.
The DMP, ProMP and OCP models perform significantly worse than the others for few kernels but become the best for more than six kernels, greatly benefiting from more kernels.
SEDS and tbGMR improve less with the use of more kernels.
Furthermore, the error distributions of DMP, ProMP and OCP become much narrower, whereas the SEDS error only marginally decreases and the spread of the error distribution remains highest.
This demonstrates that SEDS encoding can fail or be of insufficient reconstruction quality even with higher numbers of kernels. Moreover, increasing the number of kernels does not ensure convergence of the model.
For the other models, however, using more kernels significantly improves the reconstruction accuracy.
Especially the DMP, ProMP and OCP models become more and more accurate, all reaching sub-millimeter errors with $K=11$, whereas the tbGMR error is 2.6\,mm.

Considering the parameters used to encode the trajectories (see Table \ref{tab:freeparams}), the difference in using 1D or multidimensional Gaussian kernels (see Section \ref{sec:modeldifferences}) becomes apparent. Both GMM based approaches need many more parameters than DMP or OCP \textit{at any number of kernels}.
With $K=11$, DMP and OCP only need 33 parameters, whereas the tbGMR and SEDS models with $K=3$ already need 44 and 87 parameters, respectively.
ProMP without variance encoding (as only one demonstration does not define variance) uses more parameters than OCP and DMP, but not as many as the GMM based models.
When comparing the two GMM based models, the tbGMR model uses fewer parameters, because the TP-GMM extension for generalization (that nearly doubles the parameters used) is not needed for reconstruction.
Nevertheless, all encoding models compress the actual trajectory data (except for very short trajectories).
The average length of a trajectory is 83 samples, which leads to $83 \times 3 = 249$ parameters.
This number of parameters is surpassed by DMP, tbGMR, SEDS, ProMP, OCP at $K = 83, 17, 9, 20, 83$, respectively.

\subsubsection{Convergence of SEDS}
\label{sec:sedsconvergence}

As the optimization process in the SEDS model is not guaranteed to converge for all initial conditions, there were failed encodings.
As described in Section \ref{sec:seds}, we tried to mitigate this by adding random noise to the initial conditions and changing a parameter affecting stability.
Still we were not able to find an automatic way to make the optimization convergent for all trajectories in the dataset regardless of the number of kernels $K$.
With $K = 3$, 84\,\% of the optimizations converged.
The number of successful encodings increased with the higher number of kernels up to 92\,\% for $K=7$.
Out of all trajectories, only 28 could not be successfully encoded regardless of the number of kernels.
In almost all of the cases these trajectories were \textit{take out} actions from the highest target object, which are also much longer as compared to the other actions.

\subsection{Movement Generalization}

\begin{figure*}[ht!]
\centering
\begin{subfigure}{0.19\linewidth}
    \includegraphics[width=\textwidth]{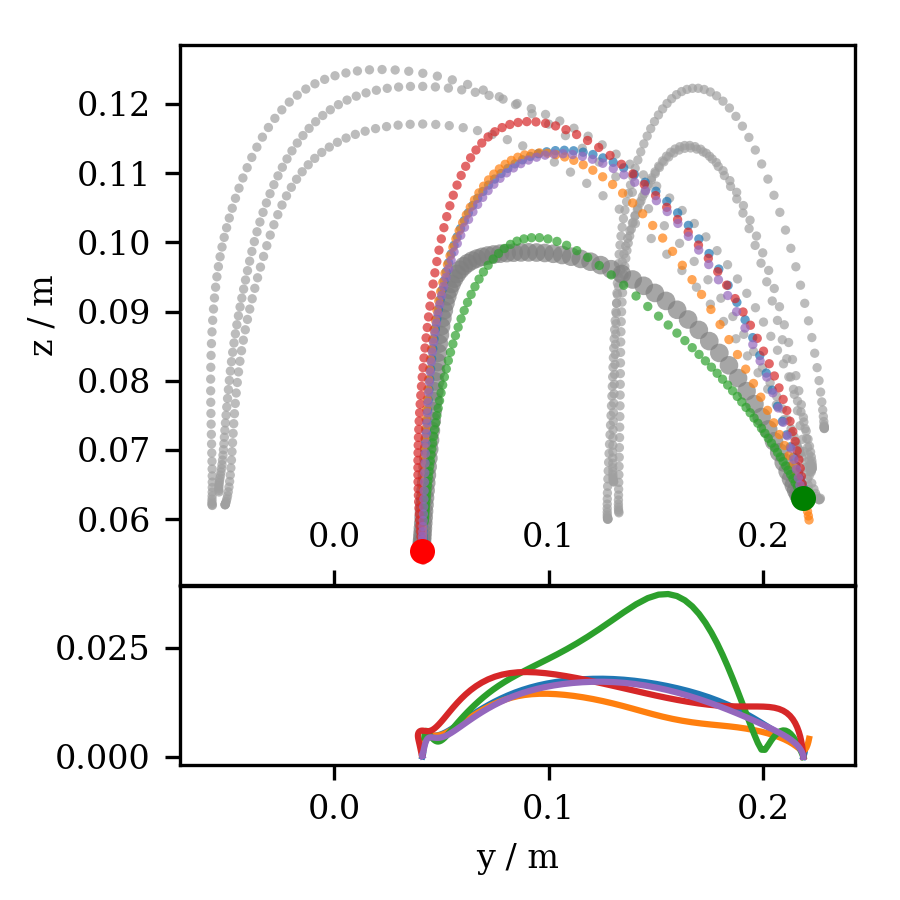}
\end{subfigure}
\begin{subfigure}{0.19\textwidth}
    \includegraphics[width=\textwidth]{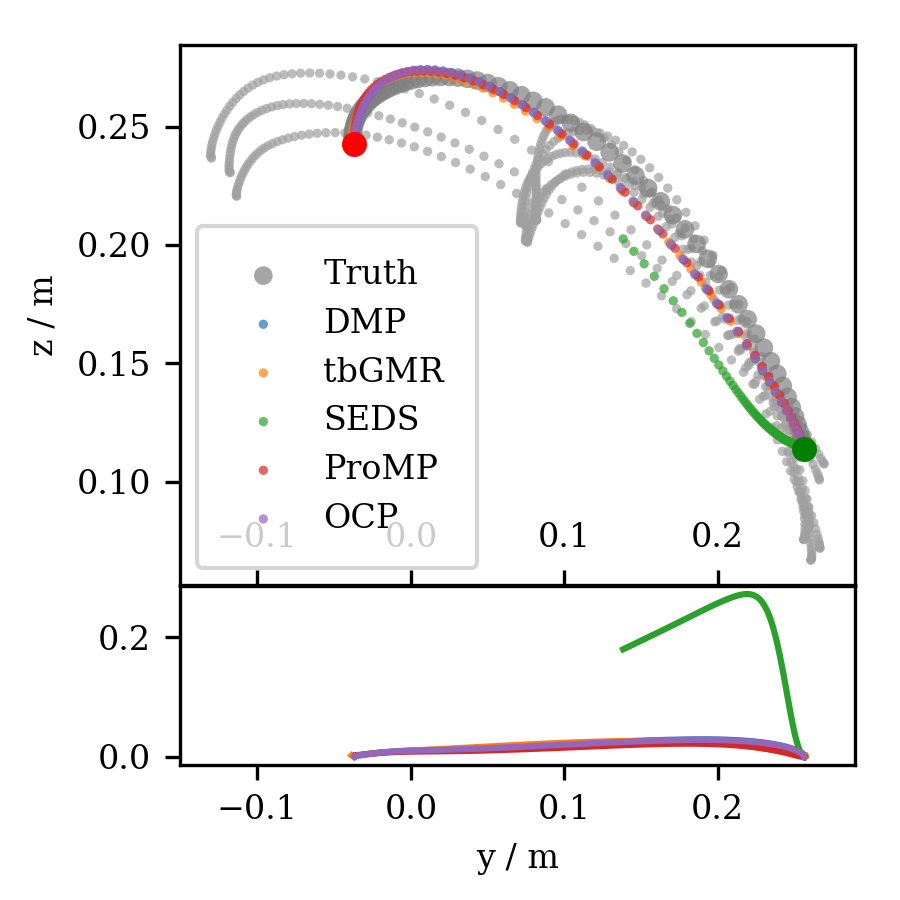}
\end{subfigure}
\begin{subfigure}{0.19\textwidth}
    \includegraphics[width=\textwidth]{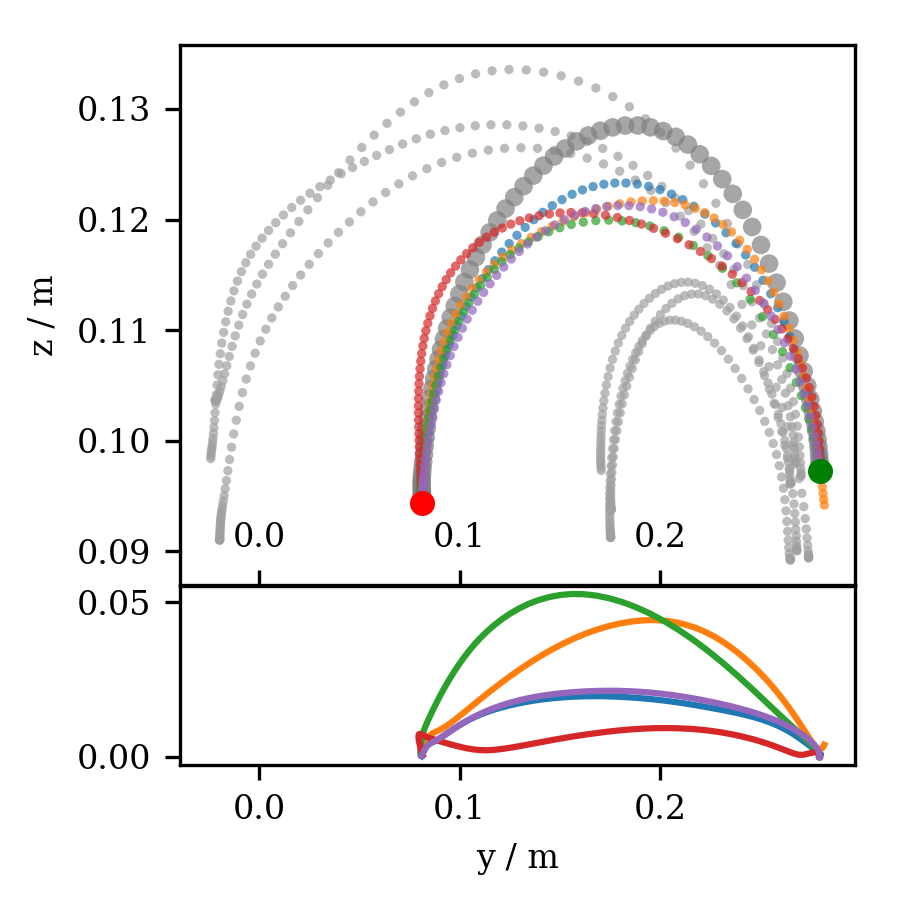}
\end{subfigure}
\begin{subfigure}{0.19\textwidth}
    \includegraphics[width=\textwidth]{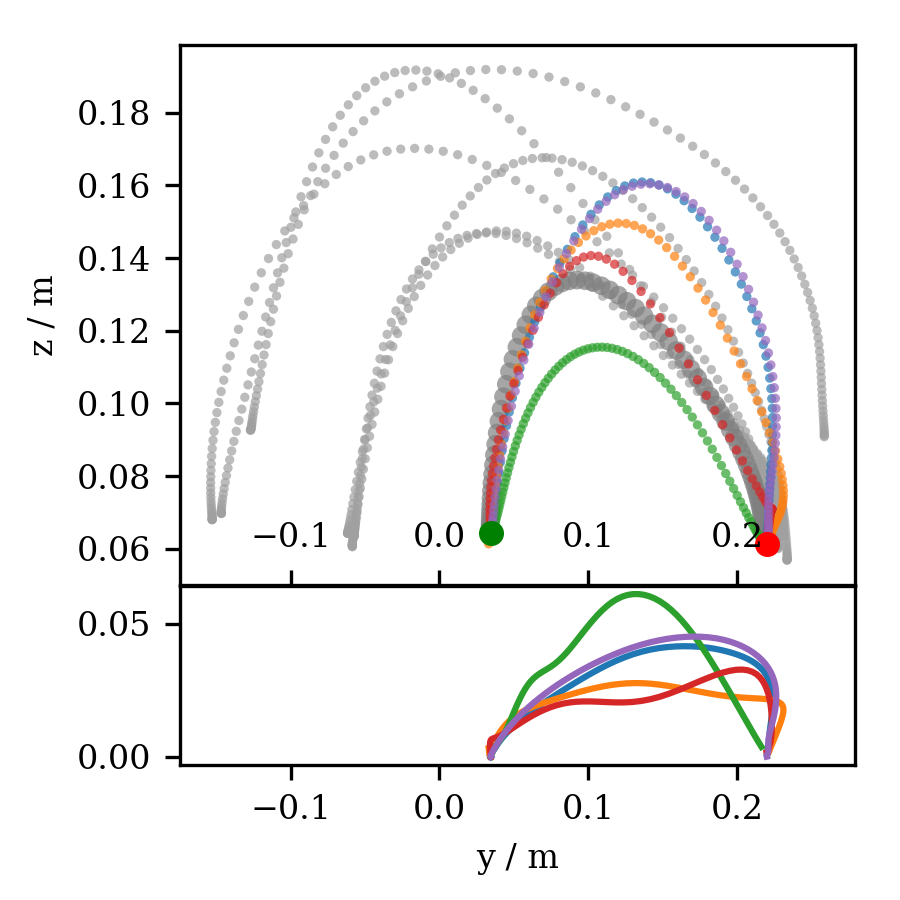}
\end{subfigure}
\begin{subfigure}{0.19\textwidth}
    \includegraphics[width=\textwidth]{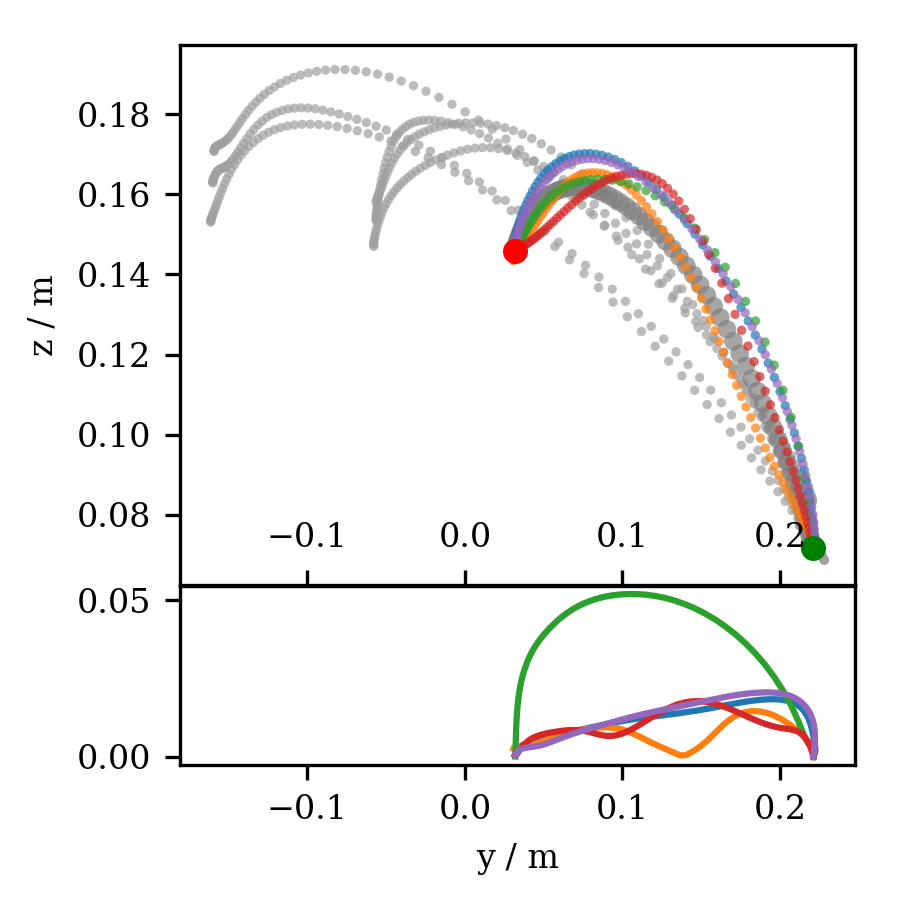}
\end{subfigure}
\captionsetup[subfigure]{oneside,margin={0.6cm,0cm}}
\begin{subfigure}{0.19\textwidth}
    \includegraphics[width=\textwidth]{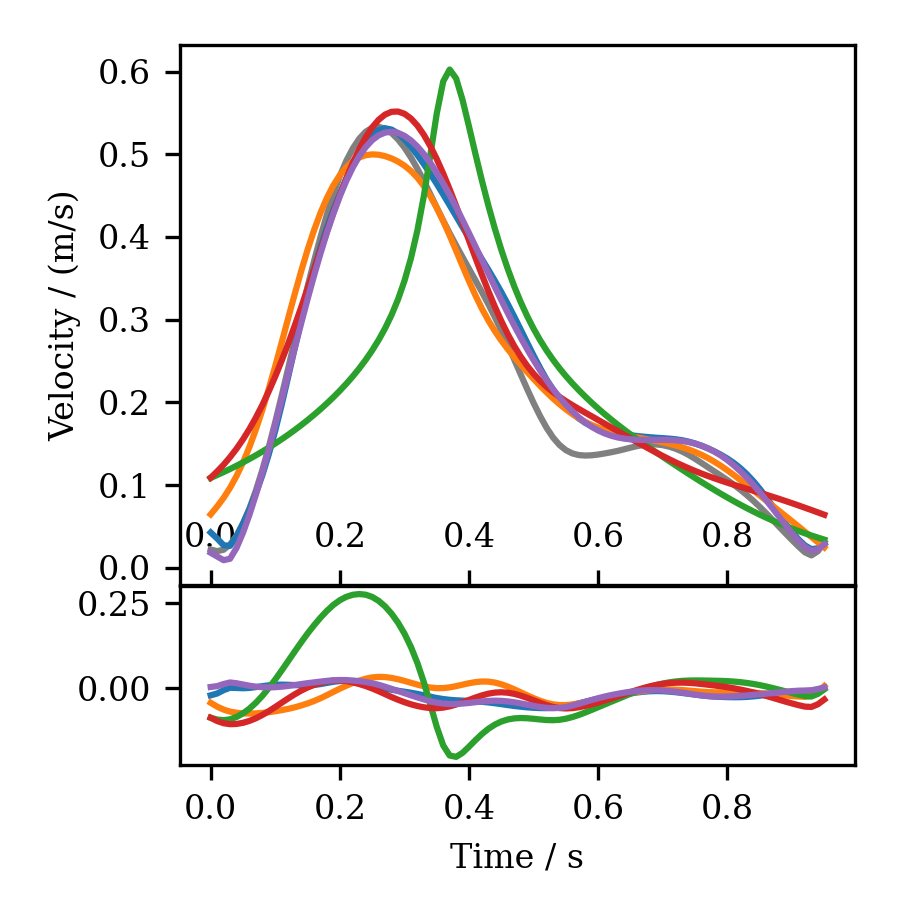}
    \caption{hide,\\interpolation}
    \label{fig:genexamples_hid_inter}
\end{subfigure}
\begin{subfigure}{0.19\textwidth}
    \includegraphics[width=\textwidth]{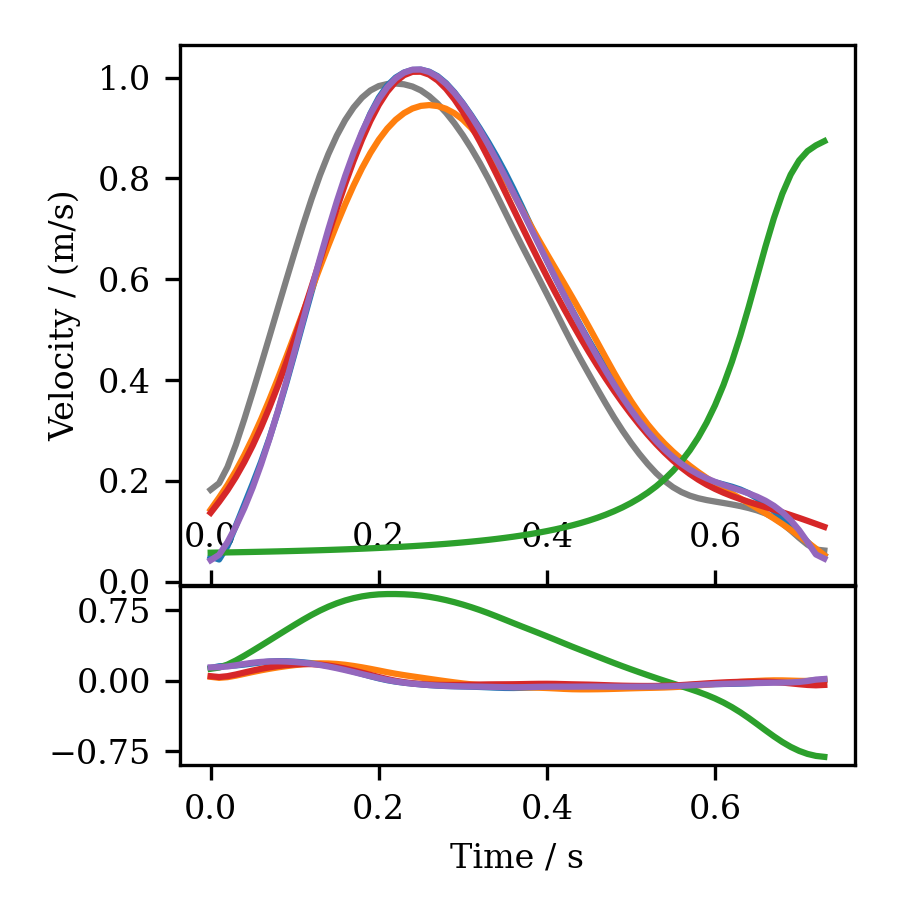}
    \caption{put on top,\\red bucket,\\interpolation}
    \label{fig:genexamples_pot-hig_inter}
\end{subfigure}
\begin{subfigure}{0.19\textwidth}
    \includegraphics[width=\textwidth]{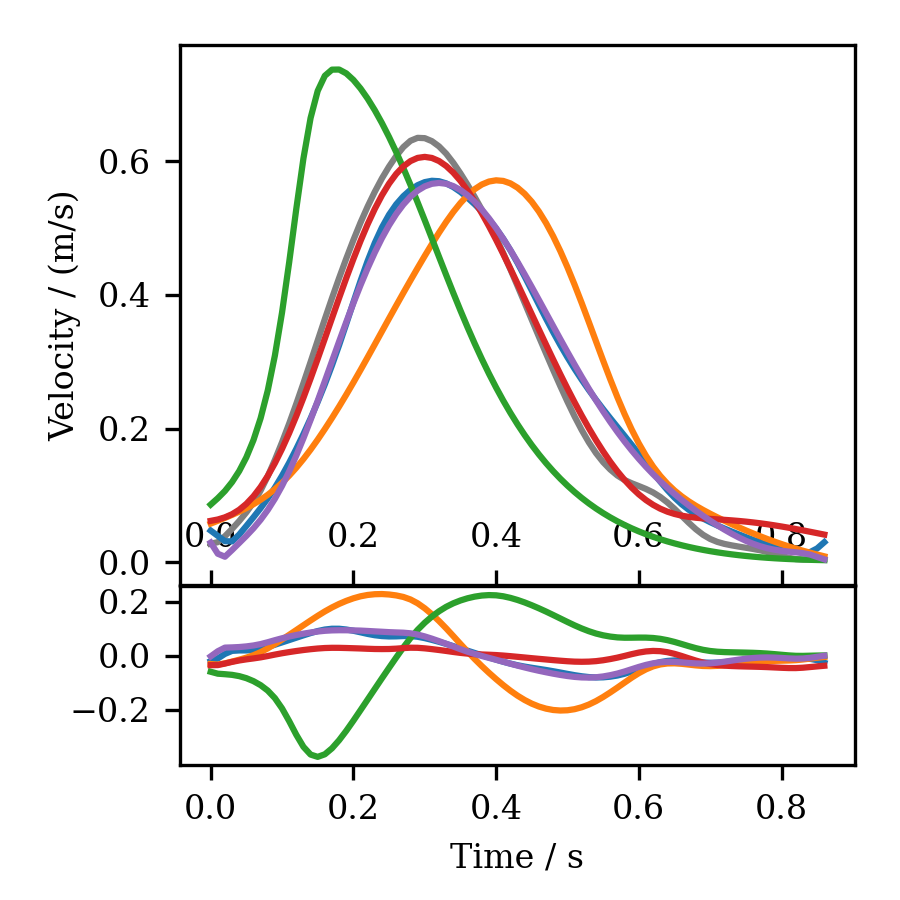}
    \caption{pick and place,\\interpolation}
    \label{fig:genexamples_pap_inter}
\end{subfigure}
\begin{subfigure}{0.19\textwidth}
    \includegraphics[width=\textwidth]{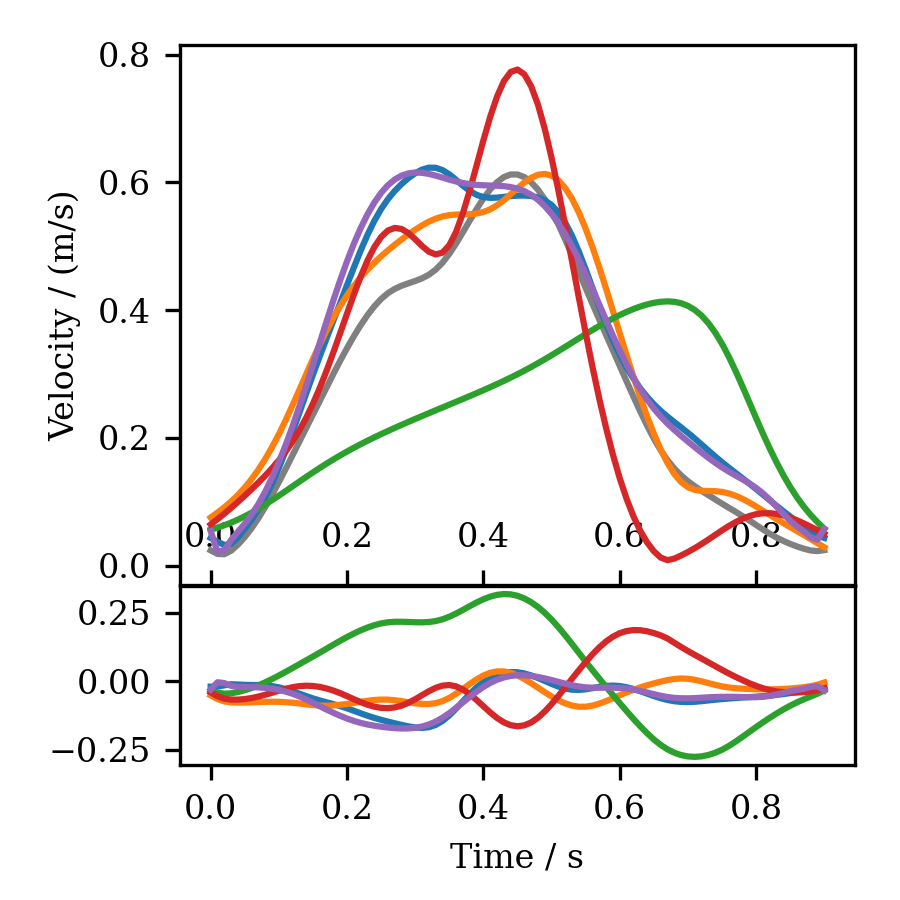}
    \caption{uncover,\\extrapolation}
    \label{fig:genexamples_hid_extra}
\end{subfigure}
\begin{subfigure}{0.19\textwidth}
    \includegraphics[width=\textwidth]{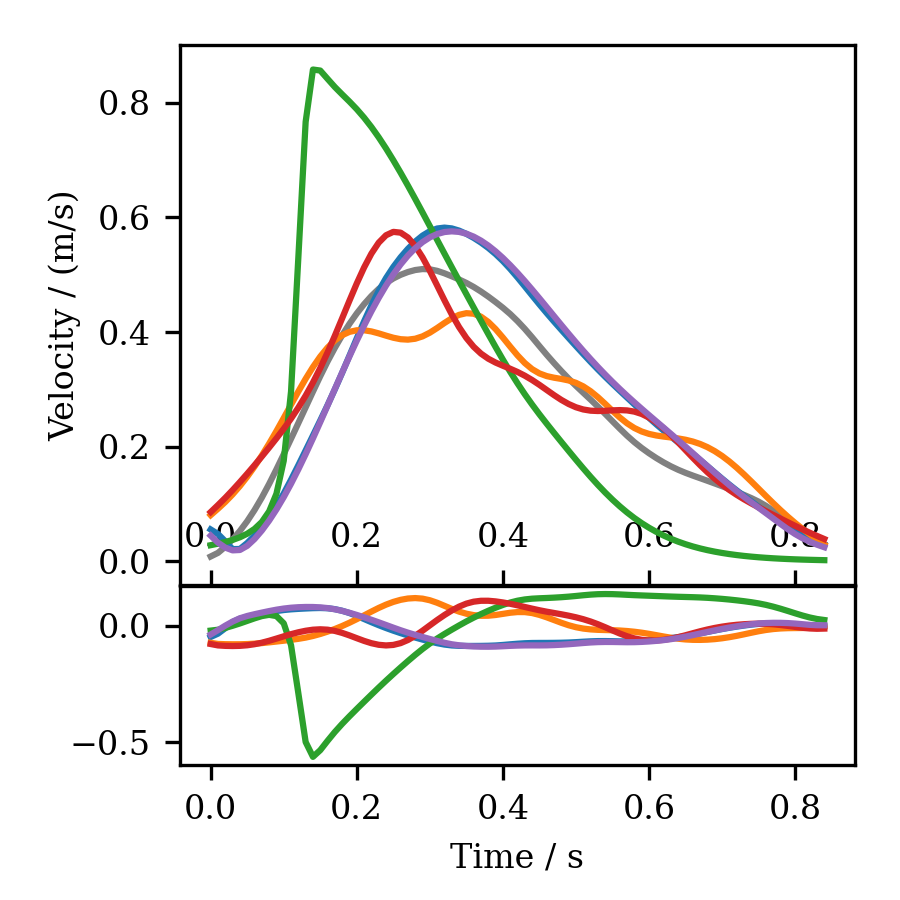}
    \caption{put on top,\\blue bowl,\\extrapolation}
    \label{fig:genexamples_pot-mid_extra}
\end{subfigure}
\caption{Examples of trajectories for different generalization situations.
         Position and velocity profiles are shown in the top and bottom rows, respectively, whereas plots at the bottom of each panel show the deviation from the human trajectory.
         Color coding for models is the same as in Figure \ref{fig:recexamples}.
         Additionally, human demonstration trajectories are shown in light gray.
         }
\label{fig:genexamples}
\end{figure*}

\subsubsection{Hyperparameter tuning}
\label{sec:hypertune_gen}

As we only recombine existing DMP and OCP encodings for generalization, hyperparameter tuning does not have to be reconsidered for these models.

However, the TP-GMM\footnote{Note that TP-GMM consists of two tbGMR models fitted on two reference frames and and not one tbGMR model as in the reconstruction case.} and ProMP hyperparameters need to be tuned again, since the fitting process is different from the reconstruction case (see Sections \ref{sec:tbgmr_generalization}, \ref{sec:promp_generalization} and Appendix \ref{app:hyperparameterspromp}).
For TP-GMMs, we observed the same qualitative behaviour as in the reconstruction case with tbGMR.
Lowering the regularization term reduces the error of position profile, but increases oscillations in the velocity profile.
To obtain smooth trajectories, the regularization term had to be chosen an order of magnitude higher than in the reconstruction case.
However, this lead on average to more than 5\,mm error at the end-point of the trajectory regardless of the number of kernels.

To circumvent this issue, we chose to use only little regularization ($10^{-6}$) and removed the oscillations by filtering the high frequencies out with a low-pass filter at 3\,Hz.
While this is a post-processing step that is not part of the model, it is neither complex nor computationally expensive.

\subsubsection{Generalisation using only few demonstration trajectories}
\label{sec:fewexamplegen}
Figure \ref{fig:genexamples} shows several examples of trajectories for movement generalization to exemplify some behavioural traits of the different models.

We observe that the SEDS model has the highest errors mainly because the velocity profile is not well matched.
Figure \ref{fig:genexamples_pot-hig_inter} shows a case where the SEDS attractor landscape is not formed appropriately.
Although at the beginning of the movement, the generalised trajectory follows the target trajectory quite well, it does it at a very low speed, and therefore it does no reach the target position in the same time as the original demonstration.

In Figure \ref{fig:genexamples_pap_inter}, we also see that the TP-GMM misses the start-point by a few millimeters.
DMP and OCP are generally very close to each other in velocity and position profile.

Furthermore, the human variability in demonstrations becomes apparent.
Especially, in Figure \ref{fig:genexamples_hid_extra} we can see that in one case the manipulation object was likely grabbed differently in one demonstration, resulting in a completely different trajectory shape with start-/end- points about 3\,cm higher than usual.

\begin{figure}[ht]
    \centering
    \begin{subfigure}{0.45\textwidth}
        \includegraphics[width=\textwidth]{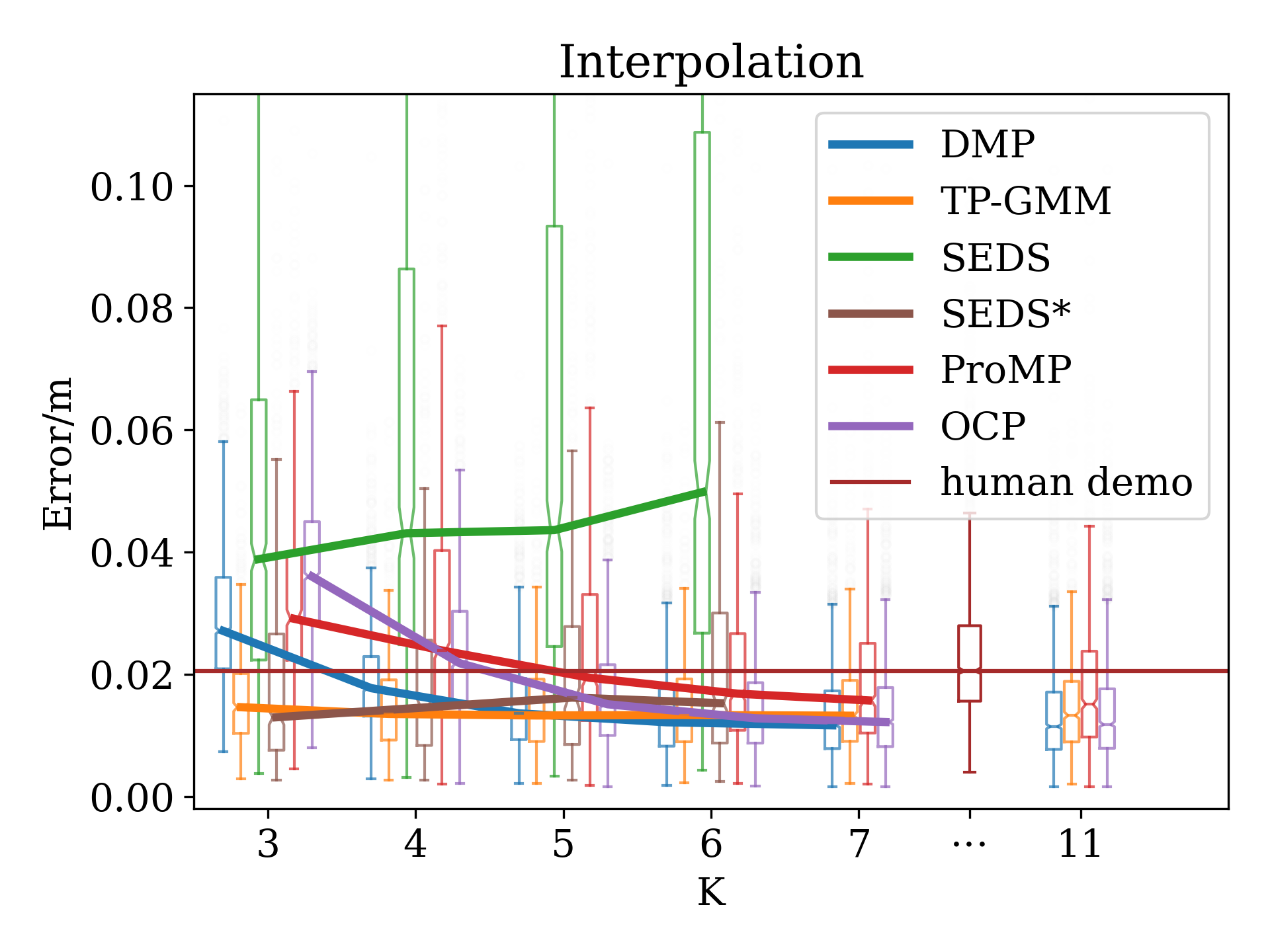}
        \caption{}
        \label{fig:fewgenerror_inter}
    \end{subfigure}
    \begin{subfigure}{0.45\textwidth}
        \includegraphics[width=\textwidth]{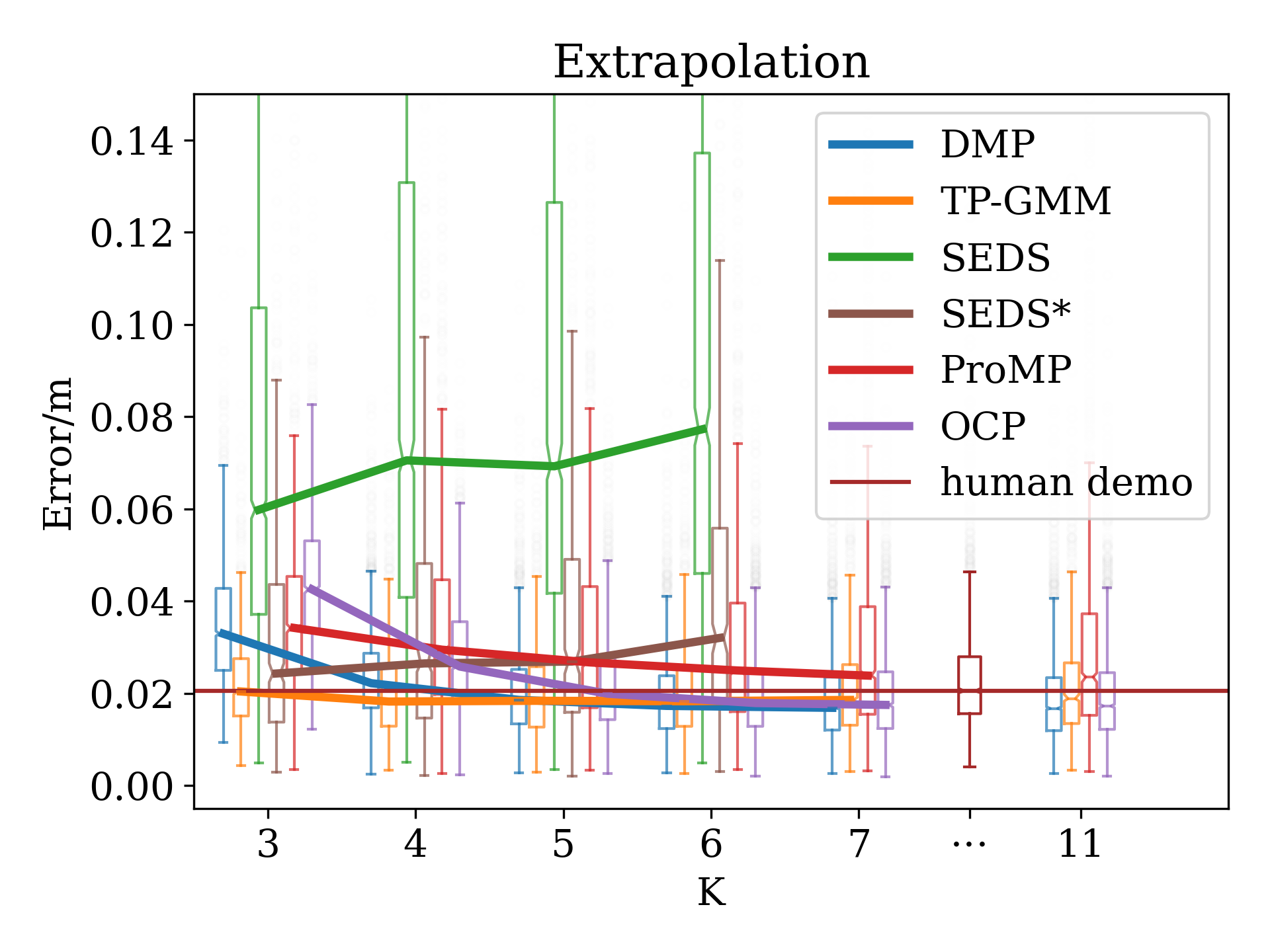} 
        \caption{}
        \label{fig:fewgenerror_extra}
    \end{subfigure}
    \caption{Comparison of movement encoding frameworks on the movement generalisation task when only few demonstration trajectories are available in case of interpolation (a) and extrapolation (b). Median error vs. number of kernels is shown for each model.
    Failed SEDS encodings are not included in this statistics (see Section \ref{sec:sedsconvergence}).
    Note, there are no results for SEDS for $K>6$ because the encoding time for high $K$ values using the repeated optimization method (see Section \ref{sec:seds_model}) becomes impractically long.
    In case of \textit{human demo}, error denotes the variance among human demonstrations whereas for models, error denotes deviation from human demonstration (see Section \ref{sec:performance_measures}).
    }
    \label{fig:fewgenerror}
\end{figure}

Statistics for the generalization error are shown in Figure \ref{fig:fewgenerror}, where the results for interpolation and  extrapolation cases are shown in (\ref{fig:fewgenerror_inter}) and  (\ref{fig:fewgenerror_extra}), respectively.
Two cases for SEDS are shown: under the condition of a fixed trajectory duration (SEDS) as for the other models and under a relaxed condition on the trajectory duration (SEDS*) as described in \mbox{Section \ref{sec:performance_measures}.}

Errors for the extrapolation cases are consistently higher than for the interpolation cases.

As in the reconstruction case, the DMP and OCP models shows a big increase in accuracy for growing $K$, but DMP needs fewer kernels than OCP to reach the same performance level.

The SEDS model performs the worst of the three models and has much wider error distributions as compared to the other two models.

In case of the relaxed time condition (SEDS*), the error is comparable to the other models for interpolation, but still higher for extrapolation as compared to DMP, OCP and TP-GMM.

The ProMP model performs better than SEDS, but worse than the other three models.
Particularly for extrapolation, its error stays higher than the human variance regardless of kernel number.

One can also observe that the errors are not converging to zero if the number of kernels $K$ is increased, but saturate for larger $K$.

Considering the number of parameters, both SEDS and TP-GMM now use nearly the same number of parameters with $29K$ and $29K-1$, respectively.
ProMP overtakes them on parameter usage at $K=7$, increasing quadratically in $K$.
Because the average number of demonstrations is 6, the DMP and OCP averaging procedure to obtain the generalised trajectory uses $18K$ parameters.
This is still less than the other models, but significantly more (six times) than in reconstruction.

\subsubsection{Generalisation using many demonstration trajectories}
\label{sec:manyexamplegen}

\begin{figure}[ht]
	\centering
	\includegraphics[width=0.9\linewidth]{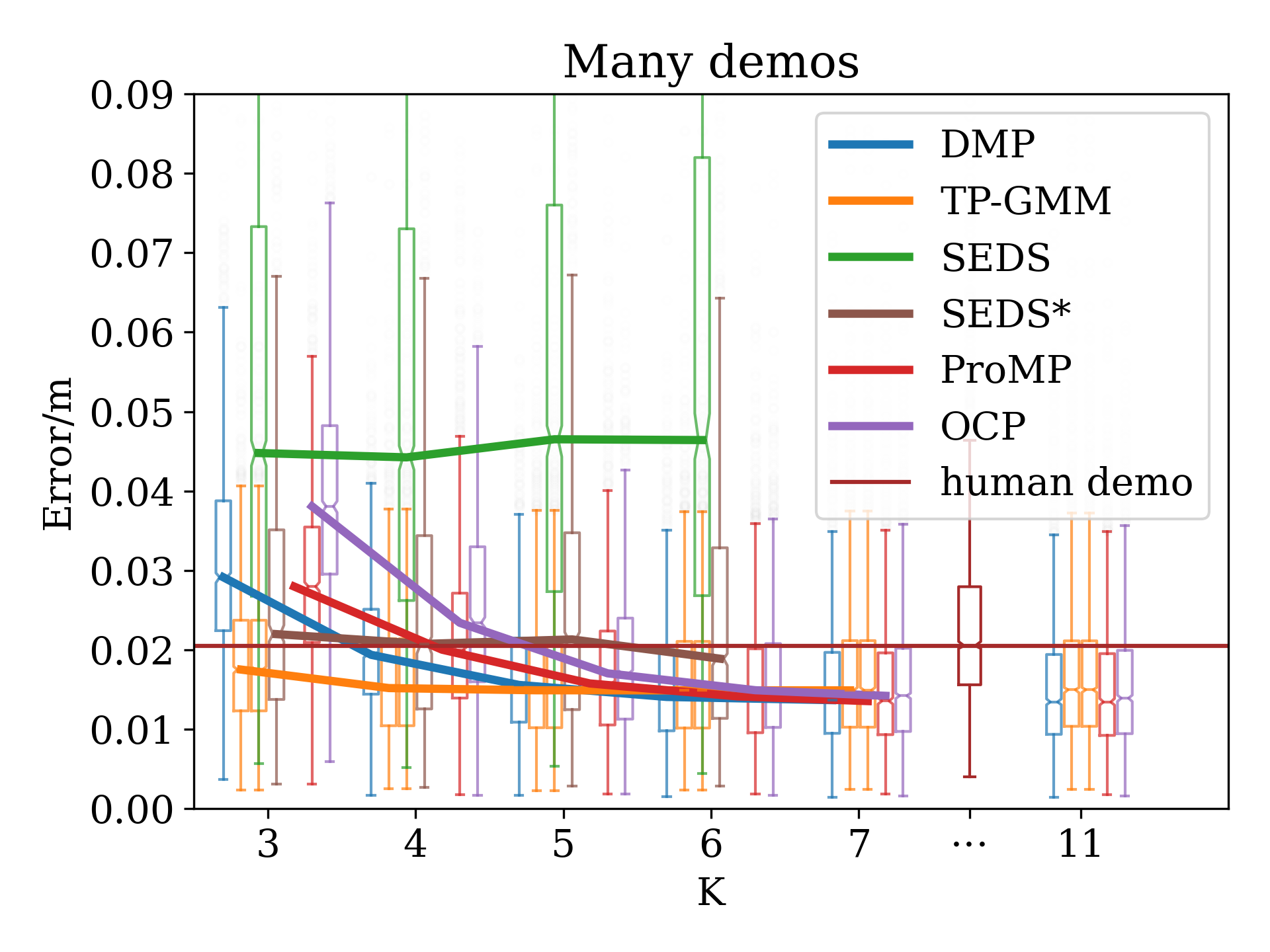}
	\caption{Comparison of movement encoding frameworks on the movement generalisation task when many demonstration trajectories are available.
	Median error vs. number of kernels is shown for each model.
    Please see the caption of Figure \ref{fig:fewgenerror} for other relevant remarks on the SEDS model.
    }
	\label{fig:manygenerror}
\end{figure}

Results for the generalisation using many demonstration trajectories are show in Figure \ref{fig:manygenerror}.
This set of generalizations contains all possible target positions with all available demonstrations, as the difference between inter- and extrapolation is not well defined in this case.
There is no qualitative change in the shape of the error curves as compared to the case of generalisation with few demonstrations.
The absolute error is comparable to the interpolation case.
The number of SEDS reconstructions with large error is smaller than with few demonstrations.
Notably, the ProMP model, which performed worse than DMP, TP-GMM and OCP with few demonstrations, performs on the same level as DMP.

SEDS, TP-GMM and ProMP still use the same number of parameters, since they encode everything into one model. Hence, the number of demonstrations has no effect on the number of parameters used.
The DMP ons OCP models, however, use more, because the set of demonstrations used is now even bigger that before.
Since there are 13 possible start-/end-points (see Figure \ref{fig:placemat}), one of which is the target, there are 12 positions in the demonstration set. With the three repetitions of each action, the average demonstration set contains 36 moves (there might be extra/forgotten repetitions or exclusions as defined in \ref{sec:generalization_procedures}).
This yields $108K$ DMP/OCP parameters that are used in the averaging procedure to obtain a generalized trajectory.

\subsubsection{Behaviour and convergence of SEDS}
\label{sec:sedsconvergenceratio}

The SEDS model gives no guarantee for end-point convergence within the specified time in case of generalised trajectories that are not similar to the demonstrations it had been fitted to. However, the smoothness of its state space leads to the reasonable assumption that it can indeed generalize to different positions.
There are many cases where the model takes much longer (see Figure \ref{fig:genexamples_pot-hig_inter}) or is faster (see Figure \ref{fig:genexamples_pot-mid_extra}) than the target trajectory.

\begin{figure}[ht]
\centering
\begin{subfigure}{0.45\linewidth}
    \includegraphics[width=\textwidth]{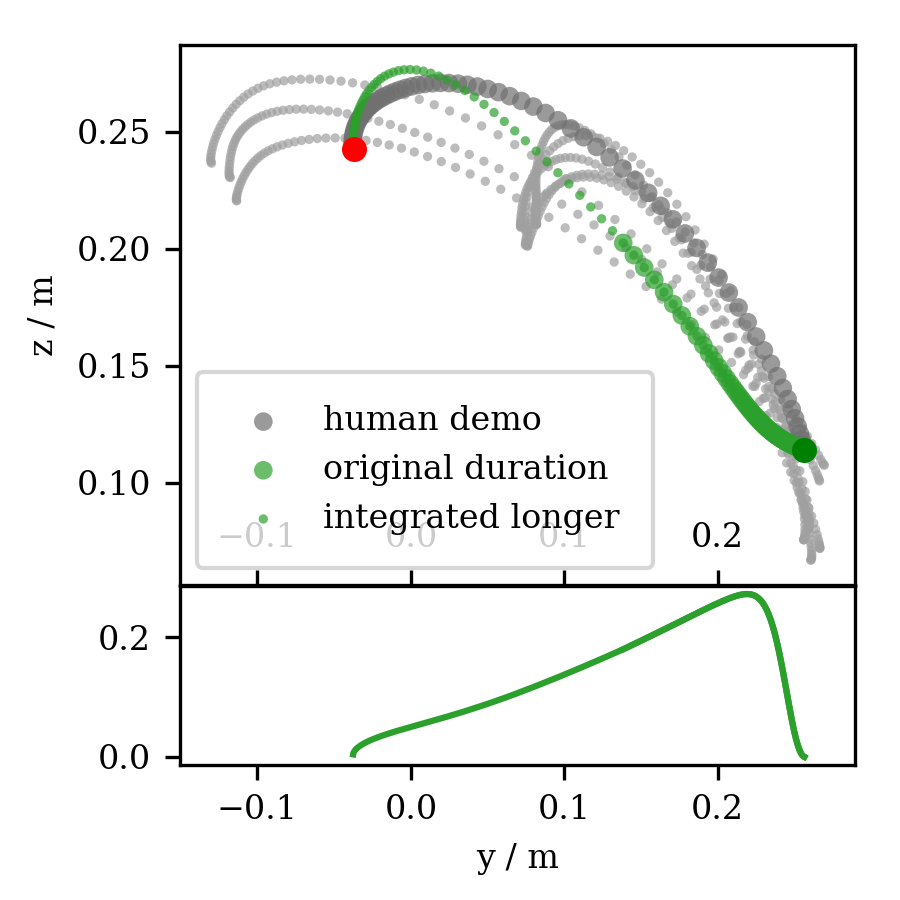}
\end{subfigure}
\begin{subfigure}{0.45\linewidth}
    \includegraphics[width=\textwidth]{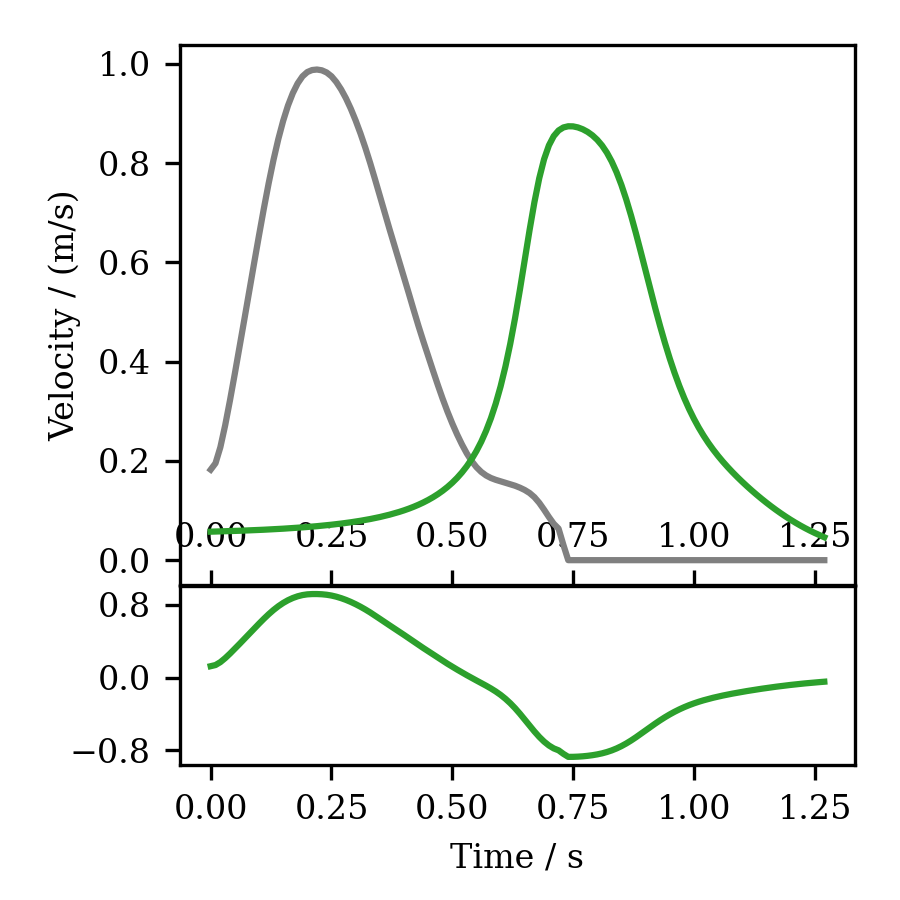} 
\end{subfigure}
    \caption{Generalization with the SEDS model for the \textit{put on top} action from Figure \ref{fig:genexamples_pot-hig_inter} when integrating the model longer than the original duration of the human demonstration.
    Position and velocity profiles are shown on the left and the right side, respectively, whereas the plots at the bottom of each panel show the deviation from human trajectory. 
    }
    \label{fig:SEDSlonger}
\end{figure}

Figure \ref{fig:SEDSlonger} shows the generalization case presented in Figure \ref{fig:genexamples_pot-hig_inter} integrated for a longer time.

This specific case does not fulfill the requirement of converging to the target on time, since it takes 1.28\,s instead of the original 0.74\,s, which is more than a 38\% increase of the movement duration.

\begin{table}[ht]
    \centering
    \caption{Percentages of successful SEDS encodings.
        \textit{Optim. success} is the percentage of cases where the optimization succeeds by the procedure defined in \ref{sec:seds_model}, regardless of the generalization quality.
        \textit{End-point conv. success} is the percentage of cases where SEDS converge to the end-point under the relaxed time condition (SEDS*) given in Section \ref{sec:performance_measures}.
         The last row contains the union over all $K$.
        The other four models have 100\% optimization success rate.
        }
    \label{tab:SEDSconvergenceratios}
    \setlength\tabcolsep{.5ex}
    \begin{tabular}{rcccccc}
        $K$ &\multicolumn{2}{c|}{Interpolation} &\multicolumn{2}{c|}{Extrapolation} &\multicolumn{2}{c}{Many demos} \\
        \hline
            &\footnotesize{Optim.}   &\footnotesize{End-point}   &\footnotesize{Optim.}   &\footnotesize{End-point}   &\footnotesize{Optim.}   &\footnotesize{End-point}\\
            &\footnotesize{success}        &\footnotesize{conv. success}      &\footnotesize{success}        &\footnotesize{conv. success}      &\footnotesize{success}        &\footnotesize{conv. success}\\
        \hline
        \hline
        3   &89\%   &58\%   &95\%   &44\%   &94\%   &51\% \\
        4   &82\%   &54\%   &96\%   &42\%   &82\%   &54\% \\
        5   &89\%   &61\%   &90\%   &39\%   &88\%   &55\% \\
        6   &78\%   &49\%   &81\%   &35\%   &71\%   &46\% \\
        \hline
        all &99.9\% &88\%   &99.7\% &73\%  &99.8\% &80\% \\
    \end{tabular}
\end{table}
Table \ref{tab:SEDSconvergenceratios} summarizes the success rates of SEDS encodings for the optimization procedure (\textit{Optim. success}) as defined in Section \ref{sec:seds_model} and for the converge to the end-point under the relaxed time condition (\textit{End-point conv. success}) as given in Section \ref{sec:performance_measures}.

We can see that in summary over all $K$ (last row of Table \ref{tab:SEDSconvergenceratios}), almost all demonstration trajectories can be encoded by the SEDS model, however, not all of those models represent generalized trajectories that converge to the target point even under the relaxed time condition.
In general, the ratio of optimization success has a tendency to decrease with higher $K$, as optimizing more kernels is harder.

\section{Discussion}

\subsection{Key Model Differences}
\label{sec:modeldifferences}

The five models investigated in  this study represent different approaches for encoding, reconstructing and generalization of movement trajectories.
They all have different requirements and give different guarantees with respect to the solutions obtained by them.
Here we briefly summarize the key differences of the models from a theoretical point of view.

DMP, SEDS and OCP are all based on a dynamical system with an attractor. Therefore, the convergence of the trajectories from any start-point to the specified end-point is always guaranteed.
Without the occurrence of perturbations, DMP and OCP ensure convergence in a given time, while with SEDS the actual duration of a generalized movement depends on the obtained model.
For ProMP and TP-GMM an additional controller is needed to provide perturbation resistance.

For the TP-GMM model, the start- and end-points can only indirectly be set by defining new task parameters.
Whether the model will reach these points or not is determined by the quality of the model fit achieved from demonstration data, however there is no guarantee that the generalised trajectory will start and end at the points specified by the task parameters.

For the ProMP model, the end points can be set exactly by conditioning the model.
However, the shape of the trajectory at the end points depends on the structure of the encoded distribution over the trajectory parameters.
If there are insufficient demonstrations around the new points, the trajectory might not look human-like.

The GMM based models and ProMP contain cross correlations between the different dimensions represented by the off-diagonal entries of the covariance matrix, whereas DMPs and OCPs have fully independent models for each dimension. While this turned out to not be beneficial for movement encoding, reconstruction, and generalisation, it might be beneficial for other tasks such as movement recognition.

In Table \ref{tab:freeparams} we provide a summary for the numbers of free parameters in the different models as derived in the previous sections.

\begin{table}[ht]
    \centering
    \caption{Number of free model parameters $n_p$ for 3D trajectories. The numbers $K$ and $N$ correspond to the number of kernels and the number of demonstrations, respectively.}
	\label{tab:freeparams}
    \setlength\tabcolsep{.5ex}
	\renewcommand{\arraystretch}{1.5}
	\begin{tabular}{lccccc}
						& DMP/OCP	& tbGMR		& TP-GMM     &SEDS  &ProMP\\
		\hline
		\hline
		Reconstruction 	& $3K$		& $15K-1$	&           & $29K$ &$\frac{1}{2}(K^2+5K)$\\
		\hline
	    Generalization 	& $3NK$		&           & $29K-1$	& $29K$ &$\frac{9}{2}(K^2+K)$\\
	\end{tabular}
\end{table}

Because the non DMP or OCP models use multivariate Gaussian kernels instead of 1D Gaussian kernels with fixed center and variance, they use significantly more parameters.
The parameter number in the covariance matrices of the GMMs and ProMP scales with $D^2$, while DMPs and OCPs only scale with $D$.

Note that the comparatively lower number of parameters in DMPs and OCPs does not directly imply that they are more efficient for trajectory compression.
While this might be true for simple reconstruction without generalization, the GMM based models and ProMPs also retain information about the variance in the demonstrations and cross correlations between dimensions, that DMPs and OCPs do not encode at all.
This information can be used by external controllers to modulate the response to perturbations based on the variance.
If the variance is high in the current situation anyway, a perturbation can be accepted and slowly corrected.
If not, quick action needs to be taken.
This can be done for OCPs as well, but only globally.
The parameters $q1$ and $q2$ define how important it is to keep the correct position or velocity, however they are not automatically adapted to the current situation.

ProMPs further have the additional benefit, that it is not necessary to pre-define possible generalization cases, like with selecting the reference frames for TP-GMM.
The model can theoretically be conditioned on any point at any time, the encoded model, and therefor the provided demonstration data, then decides if the resulting trajectory will be accurate or not.

One property of the GMM based encodings, the cross correlation between dimensions, was not assessed in this work.
More research is needed to investigate the benefit of these off-diagonal elements of covariance matrices for other tasks, e.g., movement recognition based on GMM or DMP encoding parameters.

\subsection{Influence of Hyperparameters}

For generalization with TP-GMM, it proved to be impossible to use the regularization for keeping trajectories smooth, while also maintaining good accuracy.
Especially in the case of using few demonstrations, where regularization is needed to keep the models from overfitting to the few available demonstrations, high regularization leads to higher errors and the generalized trajectories miss the target end-points in most of the cases.
Lowering the regularization reduces the error in the position profiles, however, it introduces oscillations in the velocity profiles.

The more kernels are used, the smaller the absolute value of the oscillations becomes, still the qualitative behaviour does not change.
To test if this behaviour is specific to our dataset, we reproduced the reconstruction and generalization experiments as in \cite{Calinon_tutorialtaskparameterizedmovement_2016} and looked at their velocity profiles\footnote{Note that in \cite{Calinon_tutorialtaskparameterizedmovement_2016} velocity profiles are not shown}.
We found that also with their chosen regularization, the velocity profiles of their reconstructions and generalizations oscillate as observed in our case.

Since this undesired behaviour of the TP-GMM could not be fixed by means of the model itself, we lowered the regularization to reduce errors in position profiles, and solved the velocity oscillation problem by utilising a low-pass filter.

ProMPs also exhibited oscillations when using sub-optimal hyperparameters.
For $K=3$ and $K=4$, it was even impossible to find hyperparameters that lowered the oscillations below our defined threshold (see Appendix \ref{app:hyperparameterspromp}).
However, the general reconstruction/generalization error was also high in that cases, such that the use of this model with low $K$ values becomes impractical.

\subsection{Influence of the Number of Demonstration Trajectories}

The difference between generalizing with few or many demonstrations is significant.
In case of many demonstrations, the TP-GMM model improves its end-point convergence by more than 50\% as compared to the case with few demonstrations.

The ProMP model produces worse generalizations than DMP, OCP and TP-GMM with few demonstrations.
When more demonstrations are given, however, the generalization quality reaches the same level as DMP, OCP and TP-GMM.
We suppose, the choice of modeling the distribution over the weights as a Gaussian distribution (see Section \ref{sec:promp_model}) cannot accurately represent demonstrations concentrated on only two different action classes.

For SEDS, the number of encodings with high generalization error decreases with more demonstrations.
Comparing Figures \ref{fig:fewgenerror} and \ref{fig:manygenerror}, we see that the third quartile of the error distributions when using many demonstrations is lower than for few demonstrations in most of the cases.

Furthermore, when looking at Table \ref{tab:SEDSconvergenceratios}, we see an additional dependence on nature of the presented demonstrations.
The \textit{optimization success} for individual $K$ values is higher for extrapolation than interpolation, indicating this case is easier.
But, the generalization error for extrapolation is worse than for interpolation, (also see Figure~\ref{fig:fewgenerror_inter} and Figure~\ref{fig:fewgenerror_extra}), indicating it is harder.
We suppose that \textit{encoding} an extrapolation model as such is actually easier than encoding an interpolation model, because in case of extrapolations the demonstrations are closer together than for interpolations. However, obtaining an attractor landscape for accurate generalizations outside demonstrations is less probable for extrapolation than obtaining an attractor landscape within demonstrations for interpolation, raising the number of extrapolation cases that do not meet the relaxed time convergence criteria.

For DMPs and OCPs, the usage of many demonstrations as compared to few demonstrations did not change the results significantly.
For a comparison with the simplest generalization variant with only \textit{one} demonstration, which is possible with these two models, see Appendix \ref{app:DMPGeneralization}.

\subsection{Extensions and specializations}

As discussed in the introduction, there are countless extensions and improvements to the models presented in this study.
However, all of the models but SEDS are able to produce generalizations within human variance, given enough kernels and demonstrations.
Moreover, they all end up very close to each other (see Figure \ref{fig:manygenerror} at $K = 11$).
Therefore we expect these results to be close to the optimum of what is possible given our demonstration dataset.
We do not except any specialized variant to yield substantially better results, error-wise.
However, there might be some other possible improvements regarding ease of use or additional capabilities.

The TP-GMM approach is agnostic of the underlying GMM used.
The distributions and task parameters chosen in Section \ref{sec:tbGMR} can be chosen differently, to represent the encoded trajectories in a different way.
The paper on TP-GMMs by Calinon et al. \cite{Calinon_tutorialtaskparameterizedmovement_2016} presents a lot more possibilities to encode and generalize trajectories with TP-GMMs.
Future work could test more of these encoding variants to see if any of them lead to easier hyperparameter tuning for smooth trajectories without losing precise end-point convergence.

The SEDS model presents itself as difficult to use, because of its optimization issues and bad convergence.
To improve the convergence of the SEDS model, a different solver could be used.
The paper on SEDS \cite{Khansari-Zadeh_LearningStableNonlinear_2011} mentions a custom written solver that has ``several advantages over general purpose solvers'' (see Section V. in the paper), that was unfortunately not published with the rest of the code, so we had to use the general purpose solver.
However, there is no guarantee that a better local optimum of the loss function would yield a better generalization.

Our DMP optimization algorithm could also be improved.
The implications of the very simple $\delta$-rule with its simple loss function can be clearly seen in the reconstructions with only few parameters (for example Figure \ref{fig:recexamples_few} and \ref{fig:recexamples_med}).
For $K=3$ the error is 0 at three places, the centers of the Gaussian kernels that matter for the $\delta$-rule.
To improve the encoding quality, theoretically, any other more sophisticated supervised learning rule could be used, for example one that would optimize the encoding based on the whole trajectory or adjust the centers and variances of the Gaussian kernels \cite{Ijspeert_Movementimitationnonlinear_2002,Schaal_ConstructiveIncrementalLearning_1998, vijayakumar2000locally}.
Using variable position kernels could improve the DMP results, but this would prevent the averaging of weight vectors to obtain generalized trajectories and, therefore, some other generalisation mechanism would be needed instead.
Furthermore, using biologically inspired DMPs \cite{Hoffmann_Biologicallyinspireddynamicalsystems_2009} might enable generalization from single trajectories without the problems caused by DMPs scaling properties (see Appendix \ref{app:DMPGeneralization}).

The ProMP model exhibited difficulties with generalizing in the presence of only few demonstrations.
Recently, a method which combines the TP-GMM approach (which is to a certain extent agnostic of the underlying model) with ProMPs was published \cite{Yao_ImprovedGeneralizationProbabilistic_2024}.
Using this method, to generalize in the few demonstration case, could improve the ProMP performance at the expense of having to pre-define task parameters again. 

\subsection{Encoding and Inference Time}

We will only make relative statements about the inference and encoding times of our models, since it is always possible to take more or less potent hardware with which different runt-times will be obtained.

The inference times of all three models are significantly faster then the respective encoding times. ProMPs and the GMM based models are slower than DMPs and OCPs, but still fast enough for online motion generation for robotic applications.

The differences of encoding times are more prominent.
DMPs, as the simplest model, without matrix multiplications and with a simple loss function, are significantly faster than the GMM based models.
OCPs and ProMPs are between them, but this is most definitely implementation dependent, as ProMPs also run an EM-Algorithm optimization, like the tbGMR approach.
While the tbGMR approach remains on the order of seconds, a run-time of the SEDS optimization can take on the order of a minute\footnote{On an Intel(R) Core(TM) i5-7500 CPU @ 3.40GHz with 8GB of RAM}.

\subsection{Considerations beyond our dataset}

Our dataset contains basic human manipulation actions.
However, most of the general manipulations are built up from those simple building blocks \cite{Worgotter_SimpleOntologyManipulation_2013}.
In the following we discuss how well the models may cope with more complex actions.
We cannot judge the similarity to human actions as we can with the actions in our dataset, but we can at least estimate how easy or hard it would be to adapt the models.

In more complex cases, DMPs may probably need to be augmented with via points \cite{Zhou_LearningViaPointMovement_2019} or be represented as chains of joined simpler actions \cite{Kulvicius_JoiningMovementSequences_2012}, because our simple method of averaging based on goal positions would not be applicable anymore.
As described in the introduction, there exist multiple different models for this.

OCPs would not work in more complex scenarios.
Theoretically, the LQR-controller can be used to reach any point, but the trajectory would most definitely not be human-like.
For this, a more sophisticated method to generalize the trajectory representation would be needed.
To our knowledge, no such method on a series of Chebyshev Polynomials exists.

For TP-GMMs, it would be necessary to work with more reference frames to introduce additional constraints.
However, these need to be fixed and manually designed prior to learning the models, again limiting the ability to react to any spontaneous generalization requirement.

ProMPs seem very suitable for more complex generalizations, since they can be conditioned on any via point without special preparation.
However, our results for the difference in generalization error between the few and many generalization cases indicate that this would only work well if sufficient demonstrations around the possible generalizations are provided.

Because of its global state space and smoothing constraint, it does not seem that SEDS would be able to generalize well over more complex actions.

\section{Conclusions}

In this work we compared five motions encoding frameworks including the most widely used encoding frameworks in robotics: Dynamic Movement Primitives (DMPs), time based Gaussian Mixture Regression (tbGMR) with Task Parameterized Gaussian Mixture models (TP-GMMs), stable estimator of dynamical systems (SEDS), Probabilistic Movement Primitives (ProMPs) and Optimal Control Primitives (OCPs).

We showed that the encoding of single trajectories from our dataset of human manipulation actions for reconstruction is possible for all models with an accuracy comparable to our tracking error estimate, given that enough kernels are used.
However, the SEDS model does not always converge to the demonstration just by increasing the number of kernels.

The GMM based models and ProMP with their multidimensional Gaussian kernels use significantly more parameters than DMPs and OCPs, but do not achieve better results.

Tuning the hyperparameters of the system is easiest for DMPs, OCPs and ProMPs.
When using the tbGMR approach, care has to be taken to not under-regularize the model.
While low regularization improves the reconstruction error of the position profiles, it also introduces oscillations in the velocity profiles.

Generalization from demonstrations in our dataset to new targets is possible with all analysed models, but they all have different strengths and weaknesses.
The DMP and OCP models are easy to operate, because they require no additional hyperparameter considerations, as they re-use existing encodings from reconstruction. They also produce generalizations comparable to the human inter-demonstration variation in every condition.
The TP-GMM model achieves comparable results in terms of position profiles of the generalised trajectories, but fails to reach the target end-point point precisely without additional hyperparameter considerations and post processing steps.
The ProMP model is the easiest to operate, given sufficiently many demonstrations. It is easy to regularize, requires no consideration regarding generalization (like weighting different contributions or pre-defining task parameters) and performs on the same level as the other models.
However, it yields worse results when only few demonstrations are given.

The SEDS model was not primarily developed to perform this kind of generalization, so it performs worse as compared to other models, but generalization is still possible to some extent.
However, the convergence problems with the nonlinear optimization, the inferior generalization quality and the long encoding times make SEDS impractical for many situations.

In conclusion, we showed that all models but SEDS can be used to reliably generalize to new situations.
The obtained generalizations indeed closely resemble human motion patterns.
The choice of the best model, however, depends on the application.
If only few demonstrations are available, DMP, OCP and TP-GMM can be used, with DMP needing the least parameters for sufficient accuracy, while TP-GMM should not be used if exact end-point convergence is necessary.
Given many demonstrations that cover the state space more evenly, all models but SEDS can be used, with ProMP being easiest to operate.

\section{Declaration of Conflicting Interests}
The Authors declare that there is no conflict of interest.

\section{Funding}
This work was supported by the the Volkswagen Foundation [``DeMoDiag'', grant number ZN3543] and by the German Science Foundation Grant [DFG WO 388/16-1].

\section{Research Data}
The dataset used in this work is published on Zenodo: \url{https://doi.org/10.5281/zenodo.7351664}.

\bibliographystyle{unsrt}
\bibliography{arXivVersion_3.bib}

\newpage
\appendix
\subsection{DMP Model Simplification}
\label{app:DMPSigmoid}

In this work we simplified the DMP formulation presented in \cite{Kulvicius_JoiningMovementSequences_2012} by removing the function $v$ from the computation of the function $f$.
The function $v$ is a sigmoid that drops from 1 to 0 in a very short time around the end-time $T$ of a trajectory ($v(t)=0.5$ at $t=T$).
This function was specifically designed for joining of movement trajectories which allows blending of DMP kernels at the joining point without affecting the influence of kernels at the joining point. If no joining is performed (or at the end of the joined motion), the function $v$ ensures that the trajectory converges to the goal point regardless of the shape of $f$ in time $t = T + \delta t$ (where $\delta t << T$).

\begin{figure}
    \centering
    \includegraphics[width=0.9\linewidth]{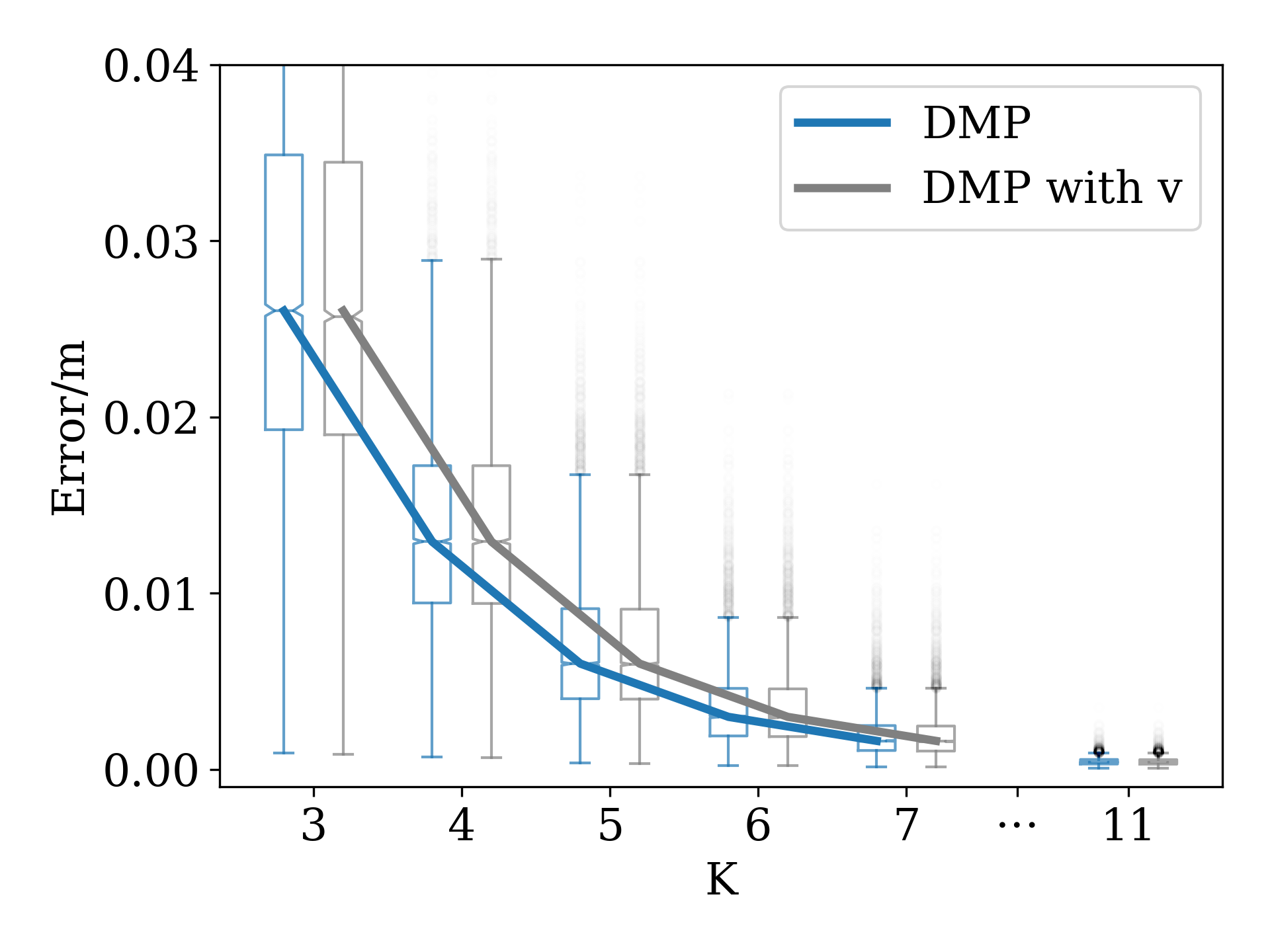}
    \caption{Comparison of DMP framework with and without $v$ on the movement reconstruction task.}
    \label{fig:dmpsigmoid}
\end{figure}

To make sure this modification does not affect the performance of the DMP formulation presented in \cite{Kulvicius_JoiningMovementSequences_2012}, we computed all DMP encodings with the sigmoid function as well.
Figure \ref{fig:dmpsigmoid} compares the reconstruction accuracy between the models with and without $v$, exactly as in Section \ref{sec:rec_accuracy}.

There is no significant difference to the DMP variant without $v$.
Since all generalizations are done using the encodings of the single trajectories, leaving out the sigmoid will have the same (no) effect on generalizations as well.

\subsection{DMP/OCP Generalization Variations}
\label{app:DMPGeneralization}
In addition to the DMP/OCP generalization method presented in  Section~\ref{sec:dmpgeneralizationmethod}, here we also include two other approaches, namely, generalization by changing the end-point, and generalisation by using weighted averaging without constraints.

\subsubsection{Generalization by changing end-point}

The simplest way to generalize a DMP/OCP to a new situation is to just change the goal point $g$ and use the weights/trajectory-representation without any changes.
Instead of averaging the available demonstrations, we select the closest trajectory with respect to the generalization target and change its goal point to the new end-point.

\subsubsection{Generalisation using weighted distance average}

Instead of computing averaging weights based on goals with constraints as explained in Section \ref{sec:dmpgeneralizationmethod}, one can also think of a simpler way to determine averaging weights.
Another way to weigh the contribution of demonstrations is just by distance to the generalization target (\cite{Weitschat_Dynamicoptimalityrealtime_2013} Section IV.A).
For this, we also translate all the trajectories to a common reference frame, but instead of obtaining the weights from expressing the new goal as a linear combination of the demonstration goals, we just weigh the trajectories by the distance $d_i$ of their goal to the new target goal.
We want close trajectories to have more influence than far away ones, so we take the inverse of $d_i$ and obtain the weights $\alpha_i$ as
\begin{align}
    x_i = 1/d_i,\\
    \alpha_i = \frac{x_i}{\sum \limits_i x_i}.
\end{align}
The second equation normalizes the weights $\alpha_i$ such that $\sum \limits_i \alpha_i = 1$.

\subsubsection{DMP performance}

We compared the generalization strategies on both the few and many demonstration tasks as defined in Section \ref{sec:generalization_procedures}.

\begin{figure}[ht]
    \centering
    \begin{subfigure}{0.9\linewidth}
        \includegraphics[width=\textwidth]{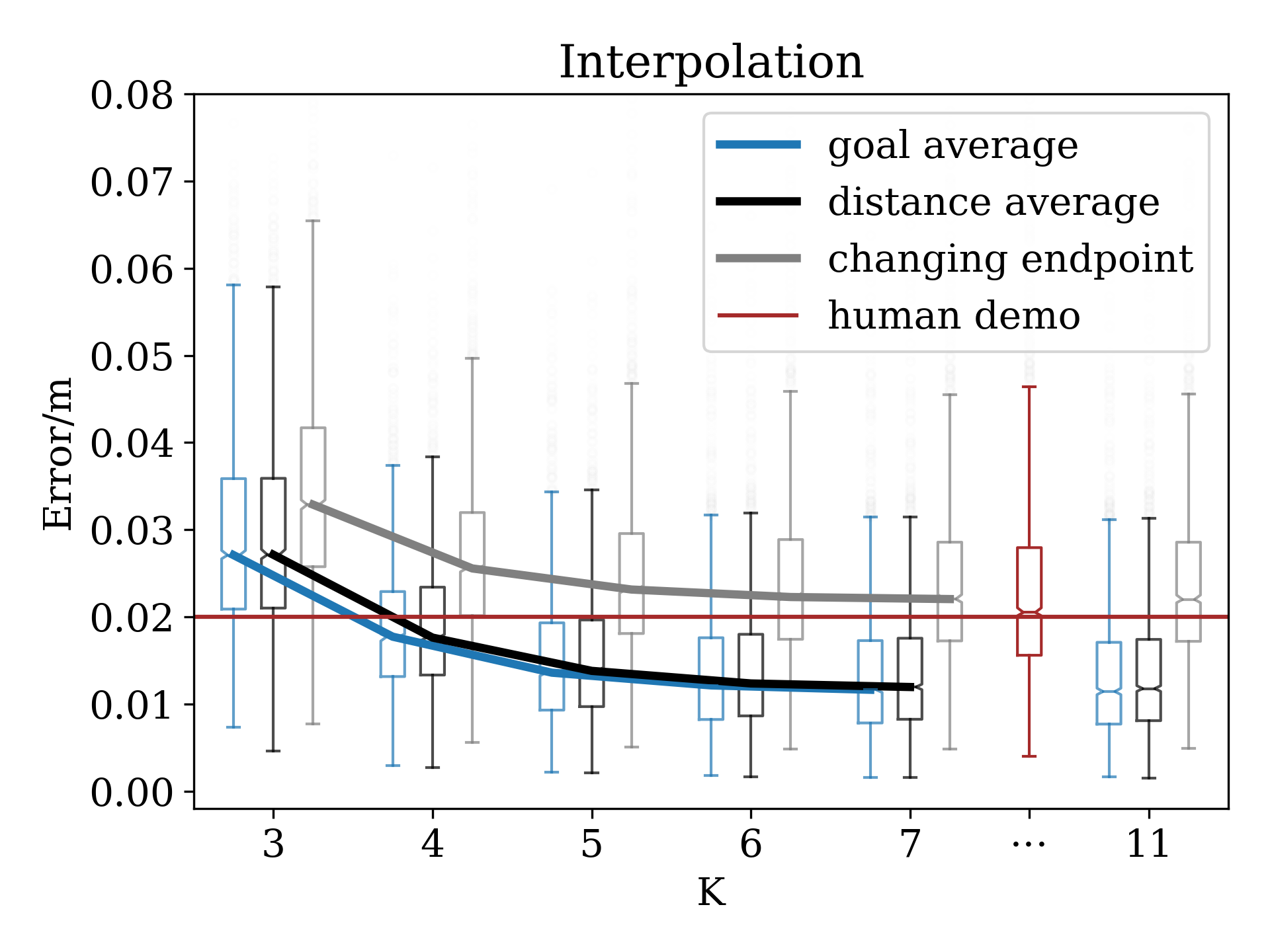}
        \caption{}
        \label{fig:dmpfewgenerror_inter}
    \end{subfigure}
    \begin{subfigure}{0.9\linewidth}
        \includegraphics[width=\textwidth]{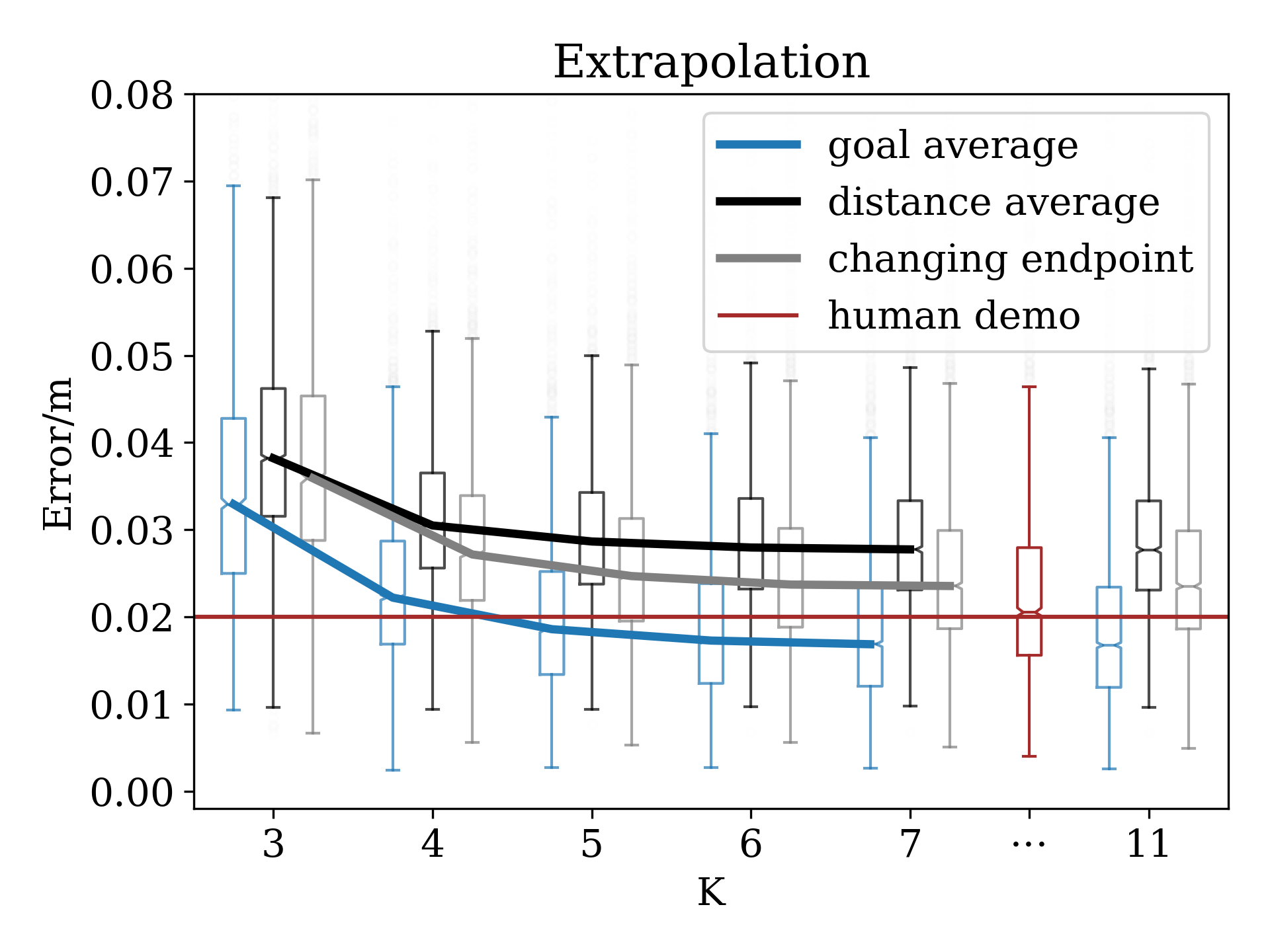} 
        \caption{}
        \label{fig:dmpfewgenerror_extra}
    \end{subfigure}
    \caption{Comparison of movement encoding frameworks on the movement generalisation task when only few demonstration           trajectories are available in case of interpolation (a) and extrapolation (b). Median error vs. number of             kernels is shown for each model.
            In case of \textit{human demo}, error denotes the variance among human demonstrations whereas for models, error denotes deviation from human demonstration (see Section \ref{sec:performance_measures}).
    }
    \label{fig:dmpfewgenerror}
\end{figure}

Figure \ref{fig:dmpfewgenerror} shows the results for the generalization error evaluated like in Section \ref{sec:fewexamplegen}.
The general trend of extrapolation errors being higher than interpolation errors is visible again.
For interpolation, goal and distance averaging perform the same, because our demonstration scenario of two bundles of demonstrations to either side of the generalization target makes them choose the same weights.
In extrapolation, there is a difference between the two, as pure distance averaging has no means of accounting for the spacial distribution of demonstrations.
So it produces DMP weights that wold be suitable for a point between the demonstrations, but closer to the ones near the generalization target.
The method is effectively still doing an interpolation, but to the wrong point.
This results in errors even  worse than simple end-point shifting.
Generalisation by changing the end-point performs worse than the inter human variance in all situations.
Additionally, there is less difference between interpolation and extrapolation than in the averaging cases.
This could be explained by the fact that selecting the closest trajectory neglects the difference between the interpolation and extrapolation scenarios.
There is always only one trajectory, so compared to averaging the spatial arrangement of the other available trajectories does not matter, leaving less difference between the two scenarios.

\begin{figure}
    \centering
    \includegraphics[width=0.9\linewidth]{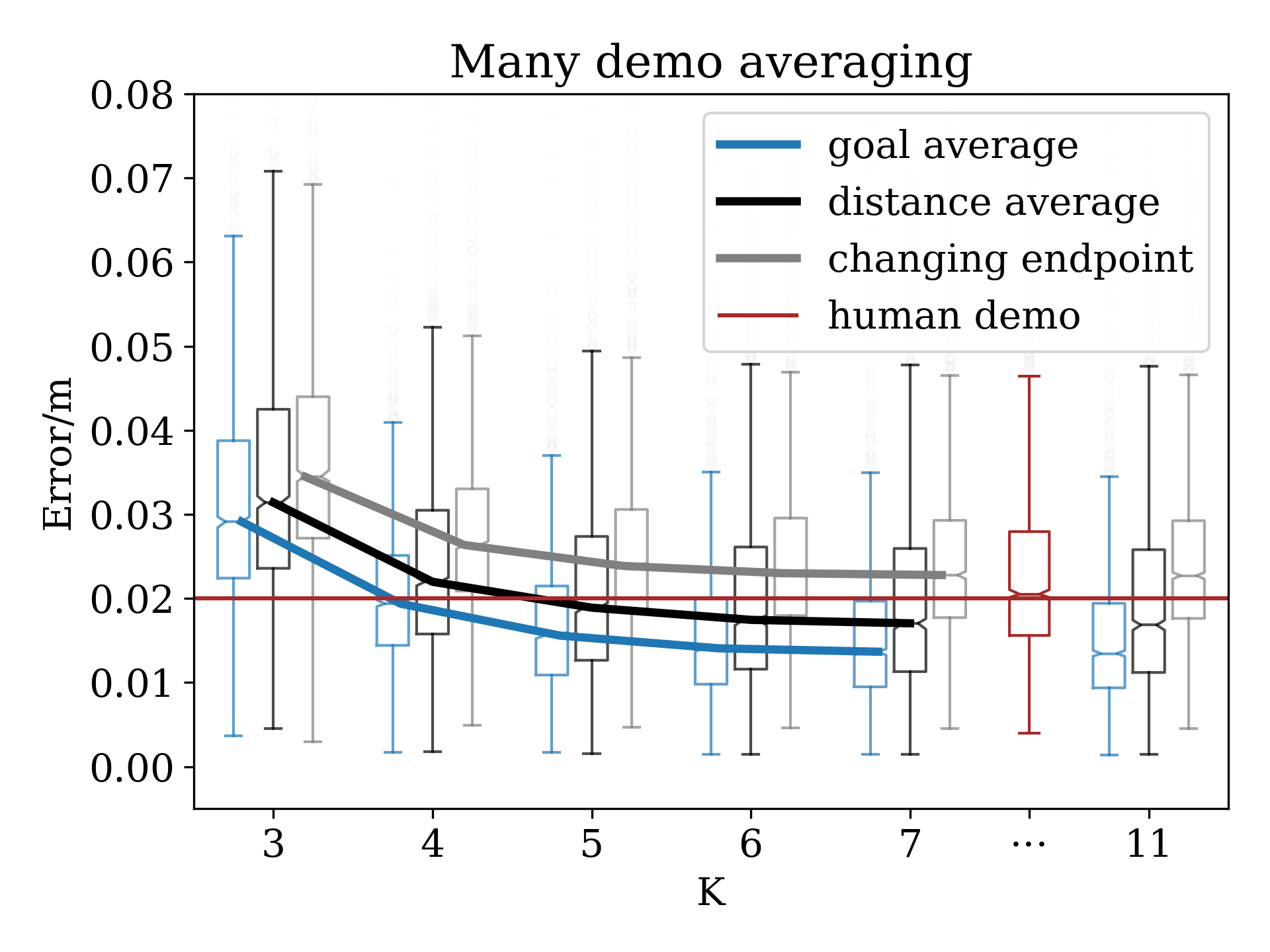}
    \caption{Comparison of movement encoding frameworks on the
            movement generalisation task when many demonstration trajecto-
            ries are available. Median error vs. number of kernels is shown for
            each model.}
    \label{fig:dmpmanygenerror}
\end{figure}

Figure \ref{fig:dmpmanygenerror}  shows the results for the generalization error evaluated like in Section \ref{sec:manyexamplegen}.
Again, goal point averaging performs the best of the three.
The simplest method of changing the end point performs worse.
Relative to the inter human human variance, both averaging methods get lower errors at high enough $K$, whereas the end-point changing stays above.

In conclusion, there is more accuracy gained by using a more complex method for obtaining new DMP weights, especially choosing an averaging approach over simple end-point changing.
On the other hand, not using averaging does not cause the model to completely fail, so the choice of method depends on demonstration availability and deviation tolerance.
However, the median error for simple end-point changing is always higher than the human variation and as the generalized trajectories really only are scaled versions of the originals, they could, while still being human-like, possibly be very large or small.
DMP modifications, like the biologically inspired ones by \cite{Hoffmann_Biologicallyinspireddynamicalsystems_2009} Hoffmann et al., have been designed to deal with this and enable better generalizations when only one demonstration is available.

\subsubsection{OCP performance}

For OCPs, the simpler generalization method of end-point changing does not work as well as for the DMPs.
It is technically possible to generalize based on one demonstration trajectory, however, practically its shape will be qualitatively different from a human demonstration.
The OCP trajectory converges to the given trajectory representation in a smooth and optimal way (as defined by the hyperparameters), as if it was recovering from a large perturbation.
This makes the generalized trajectories distinguishable from human demonstrations.
Where the DMP at least produced a human-like trajectory (as discussed above), the OCP is missing the scaling property of DMPs.
This fits well with the type of generalization showcased in \cite{Herzog_Generationmovementsboundary_2017} in Figure 7, where the end point was only slightly changed and the model performed a simple straight extension to reach the new goal.
However, this type of generalization scenario is different from the scenarios we analyzed here.

\begin{figure}[ht!]
    \centering
    \begin{subfigure}{0.49\linewidth}
        \includegraphics[width=\textwidth]{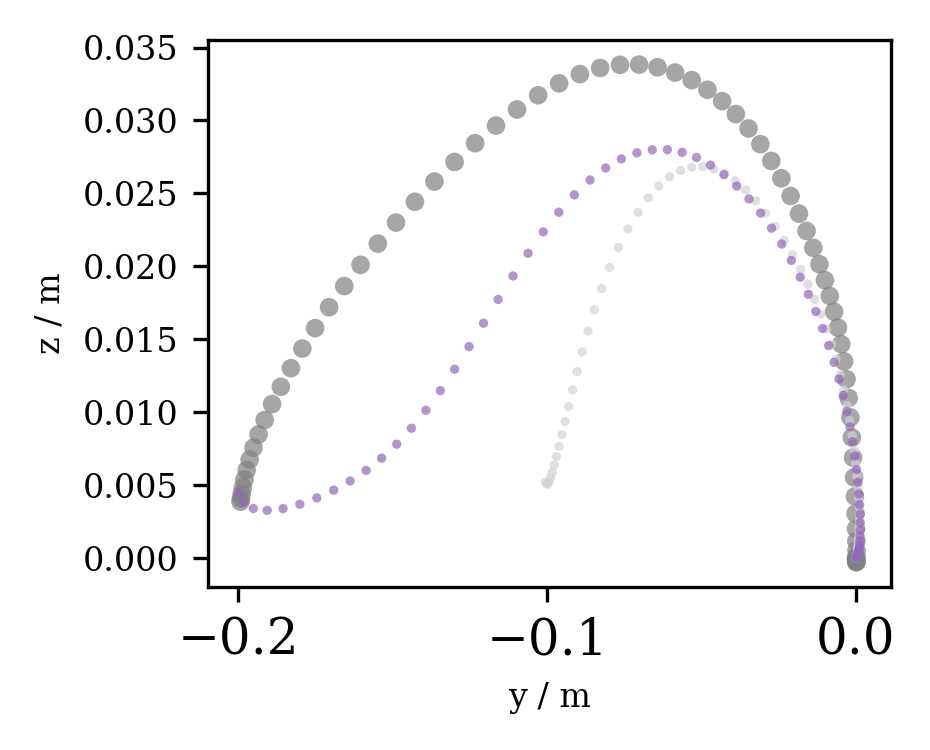}
    \end{subfigure}
    \begin{subfigure}{0.49\linewidth}
        \includegraphics[width=\textwidth]{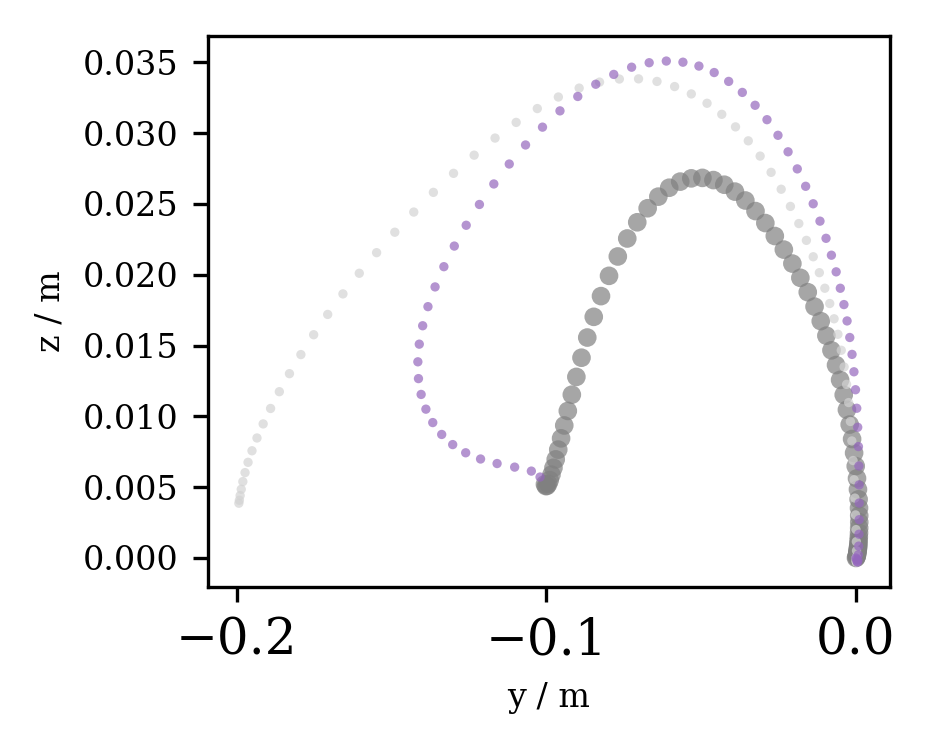} 
    \end{subfigure}
    \caption{Example OCP generalizations by end point shifting alone. The target is given in dark gray, the demonstration in light gray.}
    \label{fig:OCPsinglegen}
\end{figure}

Figure \ref{fig:OCPsinglegen} shows a demonstration of the aforementioned problem.
While the case where the demonstration trajectory was shorter than the generalization target might still be interpreted as a human action with a ``shallow'' start (which is untypical in the dataset), the one where the demonstration was larger, plainly starts in the wrong direction.

\subsection{Hyperparameter Tuning}
\label{app:hyperparameters}

All the considered models contain hyperparameters.
Choosing them optimally improves their performance.
To find the optimal hyperparameters for our models, we conducted parameter searches over sensible parameter regions on a limited set of 250 randomly chosen actions for each individual number of kernels.

\subsubsection{DMP}
\label{app:hyperparametersdmp}
To find the optimal $\kappa$, which determines the kernel variance $\sigma$ via Equation \eqref{eq:dmphyper}, we considered $\kappa$ between 0.6 and 1.2 in steps of 0.1.
$\kappa = 0.6$ corresponds to relatively narrow Gaussian kernels, where the $1\sigma$-intervals of  kernels don't overlap at all.
With $\kappa = 1.2$, the kernels are much wider such that their $1\sigma$-intervals overlap.
The optimal $\kappa$ for our data is 0.7 for $K=[3 \dots 10]$ and 1.1 for $K>10$.

\subsubsection{tbGMR}
\label{app:hyperparameterstbgmr}
Choosing the regularization parameter $\epsilon$ for the tbGMR models is more complicated than in the DMP case.
We conducted a parameter search within the range $[0.01, 10^{-7}]$.
This is from the same order of magnitude as the data, down to very little regularization that is only good for numerical stability.
While the parameter scan clearly showed that it is best to use as little regularization as possible to achieve lowest reconstruction error, the shape of those trajectories was problematic.
Different from human demonstrations, those tbGMR reconstructions looked like a piece-wise linear approximation and the velocity profile was qualitatively completely different, and contained oscillations.
Therefore, we decided to not use the low regularization for the lowest reconstruction error, because we are comparing the frameworks on human actions and those trajectories were qualitatively completely different.

We estimated the necessary regularization to keep the model's reconstructions from becoming to different from human demonstrations using two other metrics than only reconstruction error.
We observed that there were as many oscillations in the velocity profile as kernels in the model, so we compared the corresponding frequency in the FFT of the velocity profile with the fundamental frequency of the motion.
We considered a trajectory smooth enough if the oscillations caused by the kernels were lower than 20\,\% of the base frequency.

We also checked how far the end-points of the reconstructions are away for the end-points of human demonstrations. We considered 5\,mm, our estimated error from the trajectory tracking, as the maximum acceptable value.
We then selected the highest possible hyperparameters with less than 5\,mm median end-point error that still remained under our oscillation threshold.

For $K = 3$ it was not possible to fulfill both requirements at once, since the end-points were missed by more than 5\,mm regardless of regularization. Therefore, we chose to allow more oscillations in this case and used the regularization for which the end-point error was below 5\,mm.

Using the hyperparameters obtained by means of these two additional measures yields higher reconstruction errors, but the trajectories are smoother and therefore more similar to human demonstrations.

\subsubsection{SEDS}
\label{app:hyperparametersseds}

With SEDS, there are not globally optimal hyperparameters, because of the nature of its nonlinear optimization algorithm.
To deal with this, we optimize every individual fit.
We start from an estimation obtained by conventional GMM regression with the EM-algorithm, modify it to fulfill the constraints, and then refine it with SEDS optimization.
If the optimization fails, we add noise to the initial conditions and start the optimization again.
Additionally, after five failed trials, we increase the value of the \texttt{tol\_mat\_bias} value in the SEDS optimizer, that is similar to the $\epsilon$ of the TP-GMM models and is documented by the authors of the SEDS code as helping with instabilities.
We consider an optimization as failed if the numeric solver reports a fail or if the obtained model is unable to reconstruct the first given demonstration trajectory with less than 2\,cm distance anywhere along the trajectory or 5\,mm at the end point.
Note that we do not test the model against the generalization target to judge if the optimization should be repeated, which, unfairly, would use test data for training, but compare against data that was already used in training.

\subsubsection{ProMP}
\label{app:hyperparameterspromp}

ProMPs showed the same behaviour as the tbGMR approach, with oscillations in the velocity profile, directly related to the number of kernels.
The python library we used \cite{DFKIGmbHRoboticsInnovationCenter_dfkiricmovement_primitives_2021} did not expose any hyperparameters, so we modified the code to apply regularization.
Applying the same regularization as with tbGMR (additive constant in EM-Algorithm) did not change the outcome, because this only fits the distribution $p(\bm{w}, \bm{\theta})$ where the kernel properties are already set.
The library internally already had a parameter for the width of the basis functions, that was hard coded to 0.7.
We optimized this hyperparameter in the range $[0.75, 0.95]$, as the default of 0.7 was definitely to low and higher values would deteriorate the accuracy to much.

With only very few kernels (3,4), it proved impossible to get below the 20\,\% threshold as defined above for tbGMR.
But, in contrast to tbGMR, the reconstructions or generalizations with these kernel numbers were not on or below the level of human variation, so one would not use them anyway.
Thus, we accept this without attempting any further fixes.

\subsubsection{OCP}
\label{apphyperparametersocp}

Since the shape of the Chebyshev polynomials themselves is fixed and the fit of the Chebyshev series always converges, there were no hyperparameters to fit regarding the trajectory representation.
The LQR-controller itself was also easy to tune, because we do not actually need the perturbation resistance in the scenarios considered here.
So, we just made the controller as stiff as possible with a slight focus on position, rather than velocity \mbox{($q1 = 1,\;q2 = 0.1,\;R = 1$e-12)}, to follow the trajectory representation most accurately.
In actual robotic applications, this would have to be tuned depending on the specific hardware.

\end{document}